\documentclass[11pt]{article}
 
\usepackage{fullpage,url,rotating}

\usepackage{amssymb,amsmath,supertabular}
\usepackage{t1enc,amsmath,amsfonts,epsfig,graphicx,color}
\usepackage[active]{srcltx} 

\def\BibTeX{{\rm B\kern-.05em{\sc i\kern-.025em b}\kern-.08em
    T\kern-.1667em\lower.7ex\hbox{E}\kern-.125emX}}
\usepackage[active]{srcltx} 

\usepackage{bm}
\usepackage{amsmath,amssymb,amsfonts}
\usepackage{algorithmic}
\usepackage{graphicx}
\usepackage{textcomp}

\usepackage{empheq}
\usepackage{booktabs}

\newcommand{\bi}[1]{\textbf{\textit{#1}}}
\def\ST{\mathrm{ST}}

\input{def11.set}
\graphicspath{{./Figs/}{./fig/}}

\DeclareSymbolFont{YHlargesymbols}{OMX}{yhex}{m}{n}
\DeclareMathAccent{\wideparen}{\mathord}{YHlargesymbols}{"F3}

\date{}

\title{Mirror Descent and Exponentiated Gradient Algorithms Using Trace-Form Entropies}

\newcommand{\orcidA}{0000-0002-8364-7226}
\newcommand{\orcidB}{0000-0002-5056-9508}
\newcommand{\orcidC}{0000-0001-5728-0726}
\newcommand{\orcidD}{0000-0003-4121-7137}

\author{Andrzej Cichocki\footnote{Systems Research Institute of Polish Academy of Science, Newelska 6, 01-447 Warszawa, Poland,
Global Innovation Research Institute of Tokyo University of Agriculture and Technology, 2-24-16 Naka-cho, Koganei-shi, Tokyo, Japan,
Riken AIP, 103-0027 Tokyo, Nihonbashi, 1 chome-4-1, Japan; {cichockiand@gmail.com}. \orcidA{}} 
\and 
Toshihisa Tanaka\footnote{Department of Electronic and Information Engineering,
Tokyo University of Agriculture and Technology, Koganeishi,
Tokyo 184–8588, Japan; {tanakat@cc.tuat.ac.jp}. \orcidB{}} \and Frank Nielsen\footnote{Sony Computer Science Laboratories, Tokyo 141-0022, Japan; {frank.nielsen.x@gmail.com}. \orcidC{}}   
\and  Sergio Cruces\footnote{Department of Signal Processing and Communications, Universidad de Sevilla, Spain; {e-mail: sergio@us.es}. \orcidD{}}
}

\begin{document}
\maketitle

\begin{abstract}
This paper introduces a broad class of Mirror Descent (MD) and Generalized Exponentiated Gradient (GEG) algorithms derived from trace-form entropies defined via deformed logarithms. Leveraging these generalized entropies yields MD \& GEG algorithms with improved convergence behavior, robustness to vanishing and exploding gradients, and inherent adaptability to non-Euclidean geometries through mirror maps.
We establish deep connections between these methods and Amari's natural gradient, revealing a unified geometric foundation for additive, multiplicative, and natural gradient updates. Focusing on the Tsallis, Kaniadakis, Sharma--Taneja--Mittal, and Kaniadakis--Lissia--Scarfone entropy families, we show that each entropy induces a distinct Riemannian metric on the parameter space, leading to GEG algorithms that preserve the natural statistical geometry.
The tunable parameters of deformed logarithms enable adaptive geometric selection, providing enhanced robustness and convergence over classical Euclidean optimization. Overall, our framework unifies key first-order MD optimization methods under a single information-geometric perspective based on generalized Bregman divergences, where the choice of entropy determines the underlying metric and dual geometric structure.
\end{abstract}
 

\section{Introduction}

Mirror descent (MD), initially proposed by Nemirovsky and Yudin \cite{Nemirowsky}, has become an increasingly popular topic in optimization, artificial intelligence, and machine learning domains \cite{MD1,MD2,EGSD,shalev2011}.  Its profound success stems not merely from its algorithmic efficiency, but from deep mathematical connections to information geometry and the natural statistical structure underlying optimization problems. These connections, particularly to Amari's Natural Gradient (NG) method, reveal that effective optimization is fundamentally about respecting the intrinsic geometry of the parameter space rather than imposing artificial Euclidean constraints \cite{AmariNG,AmariIG,Amari09}.

The central motivation for our research emerges from a fundamental insight in information geometry: optimized  learning algorithms should adapt to the Fisher information metric of the underlying statistical manifold. This principle, pioneered by Amari in the context of neural networks, establishes that the steepest descent direction on a statistical manifold is not the Euclidean gradient, but rather the natural gradient, the direction that accounts for the curvature induced by the Fisher information matrix \cite{Amari2009,Amari-PAN}. \\
\subsection{The Information Geometry Perspective}  
The connection between Mirror Descent and Natural Gradient runs deeper than algorithmic similarity -- it represents a fundamental mathematical equivalence that has been rigorously established \cite{Raskutti2015,MD1,Amid2024}.
Our work extends this principle by showing that trace-form entropies induce natural Fisher -- like metrics that can be even more appropriate for specific problem structures. Through deformed logarithms, we can systematically explore the space of possible geometries and the possibility of discover optimal choices of hyperparameters for given data distributions  \cite{Cichocki_Cruces_Amari,CichockiAccess24}. 
\subsection{Challenge of Geometric Selection}
 While the power of geometric optimization is well established, a fundamental challenge remains: how to select the appropriate geometry for a given optimization problem? Classical approaches require domain expertise and manual tuning, limiting their applicability. Our approach addresses this through parameterized entropy families, where hyperparameters control the geometric structure \cite{MD1}.

The theoretical foundation of our approach rests on the  connection between Bregman divergences and exponential families established in information geometry \cite{AmariIG}. This connection  indicates  that choosing a generalized entropy is equivalent to selecting an appropriate exponential family structure for specific optimization problems. Our deformed logarithms allow us to systematically explore the space of possible exponential families, enabling  discovery of optimal statistical models.

The  Exponentiated Gradient (EG) and its extensions emerge as a specific and powerful instantiation of the Mirror Descent framework when the mirror map is constructed from generalized entropies and deformed logarithms. This connection is far from superficial -- it represents a fundamental mathematical relationship that unifies additive and multiplicative gradient updates within a single theoretical framework \cite{EG,KW1995,Helmbold98,CichockiAccess24,CichockiEU,CichockiTe}.

It is important to note that Mirror Descent updates can be reparameterized as Gradient Descent in appropriately chosen coordinate systems \cite{MD1,Amid2024}. This reveals that the computational complexity of Natural Gradient can be considerably reduced. Our deformed logarithms approach  gives some insight by showing that deformed logarithms naturally induce reparameterizations that preserve geometric structure while enabling efficient computation and provide implicit regularization through the choice of entropy or deformed logarithm.

\subsection{Research Contributions and Scope}

In this work, we systematically investigate the theoretical foundations and practical implications of employing trace-form entropies and deformed logarithms in the Mirror Descent framework.

Our primary contributions include:

\begin{itemize}

\item	Mathematical Framework: We establish a comprehensive mathematical foundation connecting generalized entropies, deformed logarithms, and Mirror Descent updates, providing explicit formulations for numerous  well established trace entropy families.

\item	Algorithmic Innovations: We derive novel Generalized Exponentiated Gradient  (GEG) algorithms with generalized multiplicative updates that leverage the flexibility of hyperparameter-controlled deformed logarithms, enabling  adaptation to problem geometry.

\item  The significance of this research extends beyond algorithmic development; 
It opens new avenues for understanding the geometric foundations of optimization and provides practical tools for addressing increasingly complex machine learning challenges.
The unifying theoretical framework   connects optimization theory, information geometry, statistical physics, and practical machine learning,   opens up new research directions, and provides principled approaches to algorithm design that respect the natural geometric structure of optimization problems. 

 \end{itemize}

 \section{Preliminaries: Mirror Descent (MD) and Standard Exponentiated Gradient (EG) Updates}
 {\bf Notations}. Vectors are denoted by boldface lowercase letters, e.g., $\bw \in \Real^N$, where for any vector $\bw$, we denote its $i$-th entry by $w_i$. For any vectors $\bw, \bv \in \Real^N$, we define the Hadamard product as $\bw \odot \bv = [w_1 v_1, \ldots, w_N v_N]^T$ and $\bw^{\alpha} = [w_1^{\alpha}, \ldots, w_N^{\alpha}]^T$. All operations for vectors like multiplications and additions are performed componentwise. The function of a vector is also applied for any entry of the vectors, e.g., $f(\bw) = [f(w_1),f(w_2),\ldots, f(w_N)]^T$. 
 The $N$-dimensional real vector space with nonnegative real numbers is denoted by $\Real^N_+$.
 We let $\bw (t)$ denote the weight or parameter vector as a function of time $t$. The learning process advances in iterative steps, where during step $t$ we start with the weight vector $\bw (t) = \bw_t$ and update it to a new vector $\bw(t+1) = \bw_{t+1}$. We define $[x]_+ = \max\{0,x\}$, and the gradient of a differentiable cost function as $ \nabla_{\bi w} L(\bw) = \partial L(\bw)/\partial \bw = [\partial L(\bw)/\partial w_1, \ldots, \partial L(\bw)/\partial w_N]^T $. In contrast to deformed logarithms defined later the classical  natural logarithm will be denoted by $\ln (x)$.

\subsection {\bf Problem Statement} 
We consider the constrained optimization problem:
\be
	\bw_{t+1} = \argmin_{{\bi w} \in \mathbf{R}_+^N} \left\{ L(\bw )+ \frac{1}{\eta} D_F(\bw || \bw_t)
 \right\},
	\label{Eq-1a}
\ee
where  $L(\bw)$ is a continuously differentiable loss function,  $\eta > 0$ is the learning rate and
$D_F(\bw || \bw_t)$ is the Bregman divergence induced by a strictly convex mirror map $F(\bw)$ (used here as a regularizer) \cite{Bregman1967,MD1}.
The Bregman divergence provides the geometric foundation for Mirror Descent algorithms:
\be
D_F(\bw || \bw_t) = F(\bw) - F(\bw_t) -  (\bw-\bw_t)^T f(\bw_t),
\label{Bregman1}
\ee
where the generative function $F(\bw)$ is a continuously-differentiable, strictly convex function  defined on the convex domain, while $f(\bw)= \nabla_{\bi w} F(\bw)$  is the link (strictly concave) function. 
For fundamental properties and for some extensions see, e.g., \cite{Burachik2021,Martinez2022,Nielsen2017,Nock2008}.

The Bregman divergence measures the difference between $F(\bw)$ and its first-order Taylor approximation around $\bw_t$, providing a natural measure of geometric proximity that respects the curvature induced by $F$. This geometric structure is intimately connected to information geometry—different choices of $F$ correspond to different Riemannian metrics on the parameter manifold.
The Bregman divergence $D_F(\bw || \bw_t)$  arising from generative function $F(\bw)$  referred here as mirror map can be viewed as a measure of curvature.
The Bregman divergence includes many well-known divergences commonly used
in practice, namely, the squared Euclidean distance, Kullback-Leibler divergence (relative entropy), Itakura-Saito distance, beta divergence and many more \cite{Amari2009,cichocki2010,Cichocki_Cruces_Amari,CichZd_ICA06,Cia3}.

\subsection{\bf Mirror Descent Update Rules and Geometric Interpretation}

Setting the gradient of the objective in Eq. (\ref{Eq-1a}) to zero yields the implicit update:
\be
	f(\bw_{t+1}) = f(\bw_t) - \eta \nabla_{\bi w} L(\bw_{t+1}),
	\ee
 or equivalently
\be
	\bw_{t+1} = f^{(-1)} \left[ f(\bw_t) - \eta \nabla_{\bi w} L(\bw_{t+1})\right],
\label{MDimplicit}
\ee
 where $f^{(-1)}$ is inverse function of the link function.
Note that when $F$ is separable and continuous, the inverse function $F^{-1}$ is defined globally (inverse function theorem).
In general, the implicit function theorem only guarantees local inversion of multivariate functions but not the
existence of global inverse functions.
However, when $F$ is a multivariate convex function of Legendre-type so is its convex conjugate $F^*$ and their gradients of  are reciprocal to each others globally:
$\nabla F=(\nabla F^*)^{-1}$ and $\nabla F^*=(\nabla F)^{-1}$.

Assuming that
$\nabla_{\bi w} L(\bw_{t+1}) \approx \nabla_{\bi w} L(\bw_{t})$, we obtain the explicit Mirror Descent update \cite{MD1,shalev2011}:
 \begin{empheq}[box=\fbox]{align}
\displaystyle \bw_{t+1} = f^{(-1)} \left[f(\bw_t)  - \eta \nabla_{\bi w} L(\bw_t)\right] \\
= \nabla F^{(-1)} \left( \nabla F(\bw_t) - \eta \nabla_{\bi w} L(\bw_t) \right).
		\label{f-1fDT}
\end{empheq}
 
In MD, we map our primal point $\bw$ to the dual space (through the mapping via the link function $f(\bw)=\nabla F(\bw)$) and take a step in the direction given by the gradient of the function, then we map back to the primal space by using inverse function of the  link function. The advantage of using mirror descent (MD) over gradient descent is that it takes into account the geometry of the problem through suitable choice of a link function.\\

{\bf Dual Space Interpretation}: Mirror Descent operates by 

\begin{enumerate}
\item	Mapping to dual space: $\Theta = f(\bw) = \nabla F(\bw)$,

\item	Taking gradient step: $ \Theta_{t+1} =  \Theta_t - \eta \nabla L(\bw_t)$,

\item 	Mapping back to primal: $\bw_{t+1} = f^{-1}(\Theta_{t+1})$.
\end{enumerate}
This three-step process naturally incorporates problem geometry through the choice of link function $f$.
%
%

For example, consider $F(\bw)=\sum_i \bw_i\log \bw_i$, the Shannon negative entropy.
The link function is $f(\bw) = \nabla F(\bw)=[1+\log \bw_i]_i$ with inverse map $f^{-1}(\Theta)=\left[\frac{e^{\Theta_i}}{\sum_j e^{\Theta_j}}\right]_i$.
The corresponding mirror update is the exponentiated gradient update: 
$$
\bw_{t+1}=\bw_t\odot \exp\left(-\eta\, \nabla_{\bw}\, L(\bw_t)\right).
$$
This is a standard and useful algorithm for optimization on the probability simplex that is recovered as the mirror descent with respect to the Kullback--Leibler (KL) divergence (a Bregman divergence). The underlying geometric structure is the KL Hessian geometry, an example of dually flat space in information geometry.

Note  that when the  generating function $F$ is separable across its coordinates (i.e., $F(\bw)=\sum_i F(w_i)$), 
the Hessian matrix $\nabla^2 F(\bw)$ is diagonal.

\subsection {\bf  Continuous-Time Formulation and Natural Gradient Connection}

  The continuous--time limit (as $\Delta t \rightarrow 0$) yields the mirror flow ODE:
\begin{empheq}[box=\fbox]{align}
\frac{d\,f\!\left(\bw(t)\right)}{d t}= -\mu \nabla_{\bi w} L(\bw(t)),
\label{f-1fCT}
\end{empheq}
where $\mu  >0$ is the learning rate for continuous-time learning, and $f(\bw) = \nabla F(\bw)$ is a suitably chosen link function \cite{MD1}.
Using the  chain rule, we can write mirror flow  as follows
 \be
\frac{d\,f\!\left(\bw \right)}{d t} = \frac{d\,f(\bw)}{d \bw} \odot  \frac{d\,\bw}{d t} = \diag\left\{\frac{d\,f(\bw)}{d \bw}\right\} \frac{d\,\bw}{d t}
=  -\mu \nabla_{\bi w} L(\bw(t)).
\label{chainrule}
\ee
Hence, we can obtain  continuous-time MD update in alternative form:
\be
 \frac{d\,\bw}{d t} =   -\mu \diag \left\{\left(\frac{d\,f(\bw)}{d \bw}\right)^{-1}\right\} \nabla_{\bi w} L(\bw_t)= - \mu \; [\nabla^2 F(\bw)]^{-1} \; \nabla_{\bi w} L(\bw(t))
\ee
This reveals that Mirror Descent in continuous--time is equivalent to  Natural Gradient descent with the Riemannian metric $H_F(\bw)=[\nabla^2 F(\bw)]^{-1}$. This connection, established rigorously in \cite{Raskutti2015,MD1}, shows that geometric optimization methods are fundamentally unified.

\subsection {\bf Discrete Natural Gradient Form}
 The discrete version becomes 
\begin{empheq}[box=\fbox]{align}
\bw_{t+1} = \left[\bw_t  -\eta \diag \left\{\left(
\frac{d\,f(\bw_t)}{d \bw_t}\right)^{-1}\right\} \nabla_{\bi w} L(\bw_t)\right]_+
		\label{diagMD}
\end{empheq}
where $ \displaystyle \diag \left\{ \left(\frac{d\,f(\bw)}{d \bw}\right)^{-1}\right\} = \diag \left\{ \left(\frac{d\,f(\bw)}{d w_1} \right)^{-1}, \ldots, \left(\frac{d\,f(\bw)}{d w_N}\right)^{-1} \right\} $; which we term Mirror-less Mirror Descent (MMD), representing a first-order approximation to second-order Natural Gradient methods (\ref{diagMD}) \cite{MMD2021}.
 
It should be noted that the above defined diagonal matrix can be considered as the inverse of the Hessian matrix, if it exists and has positive diagonal entries for a specific set of parameters.
The MMD is a special form of Natural gradient descent (NGD) \cite{AmariNG,Raskutti2015,MD1}. 

F. Nielsen provided a geometric interpretation of NG  and its connections with the
 Riemannian gradient, the mirror descent, and the ordinary additive gradient descent \cite{Nielsen2020}.

\subsection {\bf Canonical Examples and Some Geometric Insights}

{\bf Case 1}:  For $F(\bw) = ||\bw||^2_2/2= \frac{1}{2} \sum_{i=1}^N w_i^2$ and link function $f(\bw)=\nabla_{\bi w} F(\bw) = \bw$, we obtain the standard (additive) gradient descent
\be
		\frac{d \bw(t)}{d t}= -\mu_t \nabla_{\bi w} L(\bw(t))
\label{GDCT}
\ee
 and its time-discrete approximate version
 \be
	\bw_{t+1} = \bw_t - \eta_t \nabla_{\bi w} L(\bw_{t}).
\label{GD}
\ee

{\bf Case 2}:  For $ F(\bw) = \sum_{i=1}^N w_i \ln (w_i) - w_i$  and corresponding (componentwise) link function $f(\bw) = \ln (w)$  we obtain (multiplicative) Exponentiated Gradient (EG) update  called also as unnormalized EG update (or EGU)  \cite{EG}:
\be
		\frac{d \ln \!\left(\bw(t)\right)}{d t}= -\mu \nabla_{\bi w} L(\bw(t)), \quad \bw(t) > 0 \;\; \forall \, t.
		\label{MDEG1}
	\ee
In this sense, the unnormalized exponentiated gradient  update  (EGU) corresponds to the discrete-time version of the continuous ODE, obtained via Euler rule:
\be
	\bw_{t+1}
	&= \exp \left( \ln (\bw_t) -\mu \Delta t\, \nabla_{\bi w} L(\bw_t) \right) \nonumber \\
	&= \bw_t \odot \exp \left( - \eta \nabla_{\bi w} L(\bw_t) \right),\nonumber
\label{EG1}
\ee
where   $\odot$  and $\exp$ are componentwise multiplication and componentwise exponentiation respectively and $\eta = \mu \Delta t >0$ is the learning for discrete-time updates.
This multiplicative update naturally preserves positivity constraints and corresponds to the natural geometry of the probability simplex.

\subsection {\bf Motivation for using parameterized deformed Logarithms}
 Traditional Mirror Descent methods suffer from geometric rigidity -- the fixed choice of mirror map $F$ cannot adapt to diverse problem structures or data distributions. This limitation motivates our investigation of parameterized mirror maps based on trace-form entropies.

Adaptive Geometric Framework: Our approach addresses this fundamental limitation by introducing hyperparameter-controlled mirror maps $F_{\Theta}(\bw)$ that can:

\begin{itemize}
\item Adapt to statistical properties of training distributions.
\item Interpolate between different geometries (e.g., Euclidean, exponential family, power-law).
\item Provide automatic regularization through geometric bias.
\item 	Enable systematic geometry exploration rather than ad-hoc selection.
\end{itemize}

{\bf Information-Theoretic Foundation}: The connection between exponential families and Bregman divergences suggests that optimal mirror maps should reflect the underlying statistical structure of optimization problems. In fact, trace-form entropies provide systematic frameworks for discovering these optimal geometric structures.

There are many potential  choices of mirror map $F(\bw)$, that can  model the geometry of various optimization  problems and adapt to distribution of training data.
 In high dimensions (large scale optimization), it can be advantageous to abandon the Euclidean geometry to improve  convergence rates and performance.
Using mirror descent with an appropriately chosen function we can get a considerable improvement.

\section{Why Trace Entropies and Deformed Logarithms in MD and GEG? }

Entropy measures provide natural regularization mechanisms and geometric structures for optimization algorithms. 
The connection between entropies, information theory, and geometry runs deep; each entropy functional induces a deformed logarithm and a unique Riemannian 
manifold structure through its associated Fisher information metric \cite{naudts2002},\cite{Entropy2025},\cite{Tsallis2022},\cite{wada2010}.

Trace entropies are functionals expressible in the explicit summation form \cite{kaniadakis_scarfone2002}--\cite{kaniadakis2005},\cite{Tempesta2015},\cite{wada2023}\:
\be
S(\bp) = \sum_i p_i f(1/p_i),
\label{Gentropy}
\ee
where $p_i$ are probability values and $f(\cdot)$ is a suitable  strictly concave function. The term "trace" refers to their mathematical structure, which resembles the trace operation for matrices, i.e., a direct summation over individual components.

Trace entropies are intimately connected to deformed logarithms through  $f(x) = \log_D(x)$, where $\log_D$ represents a deformed logarithm function with specific mathematical properties ensuring proper entropic behavior \cite{kaniadakis2004},\cite{Tsallis2022}.\\
 A function $\log_D(x)$ qualifies as a deformed logarithm if it satisfies:

  \begin{itemize}

 \item

 Domain $\log_D (x)$:  $\Real^+ \rightarrow \Real$\\

 \item

 Monotonicity: $\displaystyle \frac{d \log_D(x)}{dx} >0$ \\

 \item

 Concavity:  $\displaystyle \frac{d^2 \log_D(x)}{dx^2} < 0$ \\

 \item

 Scaling and normalization: $\log_D (1)=0$, \;\;  $\displaystyle \frac{d \log_D(x)}{dx}\vert_{x=1} =1$\\

 \item

 Duality:  $\log_D(1/x) = - \log_{\tilde D}(x)$.\\

 \end{itemize}
 These axioms ensure that deformed logarithms generate well-behaved entropy functionals while providing sufficient mathematical flexibility for geometric adaptation. The concavity requirement ensures that resulting entropies satisfy the maximum entropy principle, while duality guarantees symmetric treatment of probabilities and their reciprocals, which are essential for consistent statistical interpretation.

It should be noted that deformed logarithms and their corresponding deformed exponential functions can be flexibly tuned by one or more hyperparameters, whose optimization enables the adaptation to specific data distributions and problem geometries.   By tuning/learning these hyperparameters, we can adopt to distribution of training  data and/or we can adjust them to achieve desired properties of gradient descent algorithms.

It is of great importance  to understand the mathematical structure of the generalized logarithms and its inverse generalized  exponentials in order to obtain more insight into the proposed MD or EG update schemes. Motivated by this fact,  and to make this paper more self--contained  we systematically revise fundamental properties of the deformed logarithms  and their inverses, generalized exponentials and investigate links between them.
 
We provide in the appendix the basics of $q$--algebra and $\kappa$--algebra and calculus in \S\ref{sec:qalgebra}.

 \section{{\bf MD and GEG Updates using the Tsallis Entropy and its Extensions}}

\subsection{{\bf Properties of the Tsallis $q$-Logarithm and $q$-Exponential}}

 In physics, the Tsallis entropy is a generalization of the standard Boltzmann--Gibbs entropy \cite{tsallis1988,Tsallis1994,Tsallis2022}.
 It is proportional to the expectation of the  deformed  $q$-logarithm (referred here as the Tsallis logarithm or termed logarithm) 
 of a distribution.\\
The  Tsallis $q$-logarithm is  defined for $x > 0$ as
\be
 \label{deflogq}
	\log^T_{q}(x)=\left\{
	\begin{array}{cl}
		\displaystyle \frac{x^{\,1-q}-1}{1-q}  & \text{if} \;\;  x>0 \;\; \text{and} \;\;  q \neq 1,\\
 \\
		\ln (x) & \text{if} \;\; x>0 \;\; \text{and} \;\; q=1.
	\end{array}
	\right.
\label{logq}
\ee
The inverse function of the Tsallis $q$--logarithm is the deformed $q$--exponential function $\exp^T_{q} (x)$, defined as follows:
\be
 \label{defexpq}
	\exp^T_{q}(x)=\left\{
	\begin{array}{cl}
		\displaystyle [1+ (1-q) x]_+^{1/(1-q)} & q \neq 1\\
 \\
		\exp(x) & q=1
	\end{array}
	\right.
\ee
It is easy to check that these function satisfy the following relationships:
\be
\log^T_{q}(\exp^T_{q}(x))&=&x, \;\; (0< \exp^T_{q}(x) <\infty), \;\; x> -\frac{1}{1-q} \\
 \exp^T_{q}(\log^T_{q}(x))&=&x, \;\; \text{if} \;\; x>0.
 \label{logq-prop}
\ee
{\bf Remark}: The q-deformed exponential and logarithmic functions were  introduced in Tsallis statistical physics in 1994
\cite{Tsallis1994}. However, the $q$-deformation is related to the Box--Cox transformation (for $q=1-\lambda$), which was proposed  in 1964 \cite{Box1964}.\\

\begin{figure}[htb]
	\begin{center}
		\includegraphics[width=.48\linewidth]{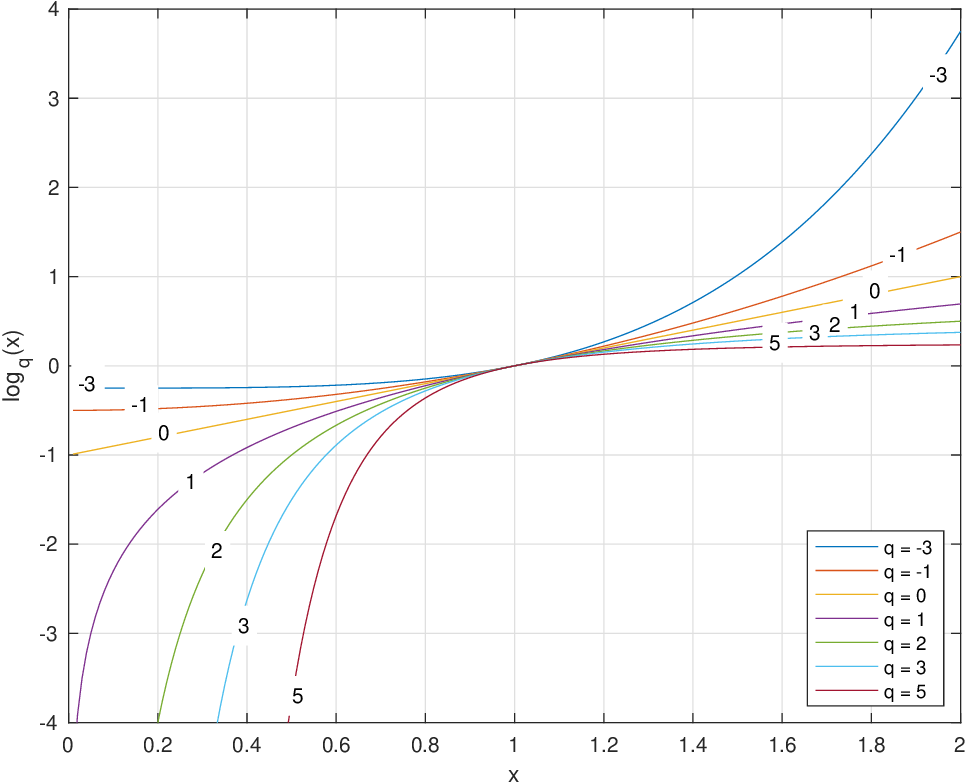}
		\hfill
		\includegraphics[width=.48\linewidth]{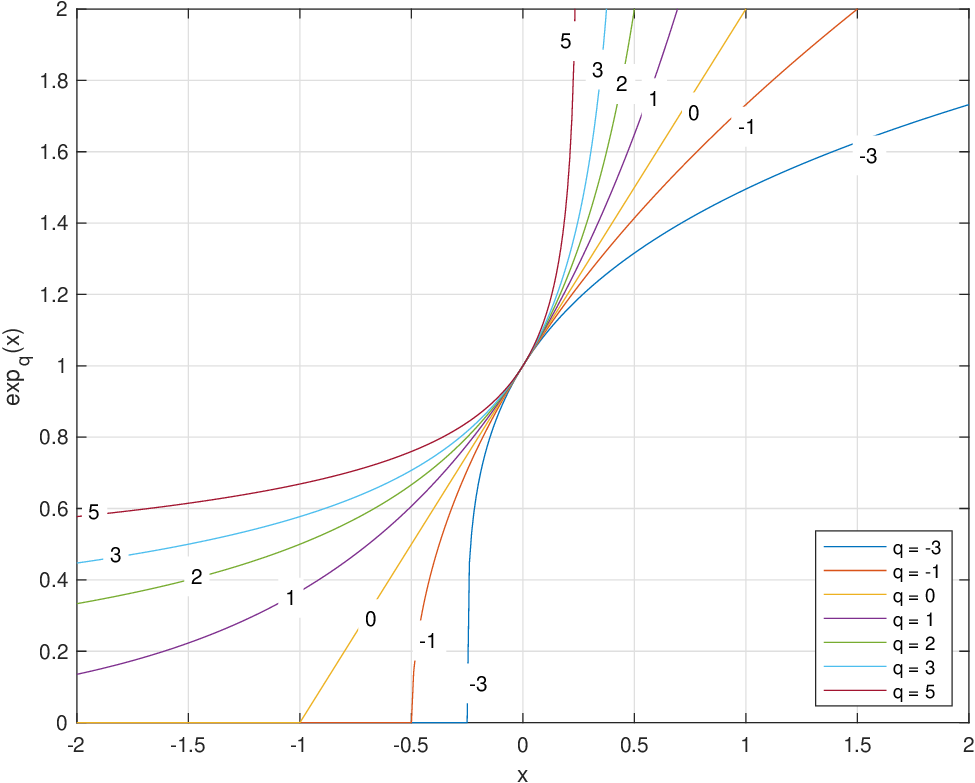}
		\caption{\color{black}Plots of the $q$-logarithm $\log_{q}(x)$ and $q$-exponential $\exp_{q}(x)$ functions for different values of parameter $q$. From the figure, one can observe how the $q$ parameter controls the degree of concavity of the $q$-logarithm as well as the degree of convexity of the $q$-exponential. Since the $q$-logarithm is convex for $q<0$, linear for $q=0$, and increasingly concave for $q>0$, particularizing to the classical logarithm for $q=1$.
		Therefore, the domain of the parameter $q$ should be limited to $(0,\infty)$ for the $q$-logarithm to satisfy the strict concavity property.
		Similarly, the $q$-exponential is concave for $q<0$, linear for $q=0$, and increasingly convex for $q>0$, particularizing to the classical exponential when $q=1$.}
		\label{Fig-def-exp}
	\end{center}
\end{figure}

The plots of the $q$-logarithm and $q$-exponential functions for various values of $q$ are illustrated in Figure \ref{Fig-def-exp}.

It should be noted that $q$-functions can be approximated by power series as follows:
\be
\log^T_q(x) \approx \ln(x) + \frac{1}{2} (|1-q|) (\ln(x))^2  +\frac{1}{6} (1-q)^2 (\ln(x))^3 +\cdots,
\ee
and
\be
\exp^T_q (x) &\approx& 1+x +\frac{1}{2} q x^2 + \frac{1}{6} \left(2q^2-q \right) x^3 \cdots \\
&=& \exp (x) +\frac{1}{2} \left(q-1\right) x^2 + \frac{1}{6} \left(2q^2-q-1\right) x^3 + O(x^4).
\ee
These functions have the  following basic properties \cite{Tsallis1994,Borges2004,Yamano2002}:
 
\be
\log^T_q(x) &=& - \log^T_{2-q} (1/x)\\
\frac{\partial \log^T_{q} (x)}{\partial x}  &=& \frac{1}{x^q}  > 0\\
\frac{\partial^2 \log^T_{q} (x)}{\partial x^2}  &=& \frac{-q}{x^{q+1}} <0  \;\;\text{for}  \;\; q >0.
\label{expq-prop}
\ee
It is easy to prove the following fundamental  properties:
\be
\log^T_{q}(x y) =\log^T_{q}(x) + \log^T_{q}(y) + (1-q) \log^T_{q}(x))  \log^T_{q}(y) \;\; \text{if} \;\; x>0, y>0
\label{logxy}
\ee
\be
\exp^T_{q}(x) \exp^T_{q}(y) =\exp^T_{q}(x+y +(1-q) x y).
\label{expxexpy}
\ee
Using these properties we can define nonlinear generalized algebraic forms
$q$-sum  and the $q$-product (see for more details about q-algebra in \cite{Borges1998,Borges2004}
\be
x \oplus_q^T y  &=& x +y +(1-q) x y, \qquad (x \oplus_1^T y =x+y),\\
x \otimes_q y &=& [ x^{1-q} + y^{1-q} -1]_+^{1/(1-q)} \;\; \text{if} \;\; x>0, \;\; y>0 \qquad (x \otimes_1^T y=  x \; y).
\label{q-prod}
\ee
Using this  notation, we can write the following formulas

\begin{empheq}[box=\fbox]{align}
\exp^T_{q}(x+y) =\exp^T_{q}(x) \otimes_q   \exp^T_{q}(y),\\
\exp^T_{q}(\log_q^T (x) +y)= x \otimes_q   \exp^T_{q}(y),
\label{expx+y}
\end{empheq}

which play a key  role in this paper.

\subsection{{\bf MD and GEG Updates Using the Tsallis $q$-logarithm}}

Let assume that the link function in Mirror Descent can take the following componentwise form
\be
f_q(\bw) = \log^T_q (\bw),\qquad \bw=[w_1, \ldots, w_N]^T \in \Real_+^N.
\label{linkfq}
\ee
In this case the mirror map $F(\bw)= \sum_i w_i \log_q (w_i) - \log_{q-1} (w_i)$  and the Bregman divergence is a well known beta divergence \cite{Cia3}:
\be
D_{F_q}(\bw_{t+1}|| \bw_{t}) &=&  \sum_{i=1}^{N}  w_{i,t+1} \left(\log_q(w_{i,t+1}) 
-\log_q (w_{i,t}) \right)
- \log_{q-1} (w_{i,t+1}) + \log_{q-1} (w_{i,t}) \nonumber \\
&=&  \sum_{i=1}^N w_{i,t+1} \frac{w_{i,t+1}^{1-q} -w_{i,t}^{1-q}}{1-q} - \frac{w_{i,t+1}^{2-q} -w_{i,t}^{2-q}}{2-q} ,
\label{Bregmanq}
\ee
where $\beta=1-q, \;\: \beta \neq  0$.

Applying Eq. (\ref{f-1fDT})  and taking into account formula (\ref{expx+y}), we obtain novel generalized exponentiated gradient update referred to as  $q$-GEG or $q$-MD update
\begin{empheq}[box=\fbox]{align}
\bw_{t+1} =  \exp_q^T \left[ \log_q^T (\bw_t) \otimes_q   \left(-\eta \nabla_{\bi w} L(\bw_t) \right) \right]=  \bw_t \otimes_q  \exp_q^T \left(-\eta \nabla_{\bi w} L(\bw_t) \right)
\label{qEG1}
\end{empheq}

where $ \otimes_q $ $q$-product  defined by Eq. (\ref{q-prod}) is performed  componentwise.

The above $q$-MD update can be written in a scalar (componentwise) form as
\begin{empheq}[box=\fbox]{align}
w_{i,t+1} = w_{i,t}  \otimes_q  \exp_q^T \left(-\eta \nabla_{w_{i}} L(\bw_t) \right) =  \nonumber \\
= \left[ w_{i,t}^{1-q} + \left(\exp_q^T(-\eta \nabla_{w_{i}} L(\bw_t))\right)^{1-q} - 1 \right ]_+^{1/1-q}
\label{qEGscalar}
\end{empheq}

By choosing  e.g., $\alpha =3$, we can   alleviate the  well-known problem occurring in the GD updates of vanishing and exploding gradients.

By exploiting fundamental properties of the Tsallis exponential function, we can easily prove the identity

\be 
\exp^T_q(x+y) = \exp^T_q (x) \;  \exp_q^T \left(\frac{y}{1+(1-q)x} \right).
\label{expx+y2}
\ee

Hence, we obtain simplified generalized $q$-GEG  update
\begin{empheq}[box=\fbox]{align}
w_{i,t+1} = w_{i,t}  \exp_q^T \left(\frac{-\eta \nabla_{w_{i}} L(\bw_t)}{w^{1-q}_{i,t}} \right),
\label{qEG2scalar}
\end{empheq}

which can be written in a compact vector form 
\begin{empheq}[box=\fbox]{align}
\bw_{t+1} = \bw_{t} \odot \exp_q^T \left[- \eta_t \odot \nabla_{\bi w} L(\bw_t)\right]
\label{qEG2vect}
\end{empheq}

where a vector of learning rates
 $$
\eta_t =[\eta_{1,t},\ldots,\eta_{N,t}]^T,
$$ 
has entries  
$\eta_{i,t}= \eta/\left((1+(1-q)\log_q^T(w_{i,t})\right)= \eta \; w^{q-1}_{i,t}$.\\
{\bf Remark}:  Assuming that learning rate is time variable  and it is represented by a vector i.e., $\bm \eta \rightarrow \eta \bw_t^{1-\alpha}$, we obtain $\eta_t= \eta \bw_t^{1- \alpha- \beta}$ and derived update takes particular form derived and extensively experimentally 
tested in our recent publication \cite{CichockiAccess24} (by exploiting 
alpha-beta divergences \cite{cichocki2010,Cichocki_Cruces_Amari}).

\subsection{\bf {MD  and EG  Using  Schw{\"a}mmle-Tsallis (ST) Entropy}}

Schw{\"a}mmle and Tsallis proposed two parameter entropy \cite{schwammle2007}
\be
S_{q,q'}^{ST} (p) = \sum_{i=1}^{W} p_i \log_{q,q'}^{ST}(1/p_i),\;\;
q\neq 1, \;\; q'\neq 1,
\label{STentropy1}
\ee
where the deformed logarithm referred to as the ST-logarithm or ST $(q,q')$-logarithm is defined as
\be
\log_{q,q'}^{ST}(x) = \log^T_{q'}([x]_q) =\log_{q'}^T e^{\log_q^T (x)} = \frac{1}{1-q'} \left[\exp\left(\frac{1-q'}{1-q}(x^{1-q}-1)\right)-1\right]
\label{STlog}
\ee
for $x>0$ and $q \neq q'$ (typically $q>1$ and $q'<1$), where $[x]_q= \exp(log_q(x))$ and its inverse function is formulated as a two-parameter deformed exponential, $\exp_{q,q'} ( x)$
\be
\exp_{q,q'}^{ST} (x) = \left[ 1+ \frac{1-q}{1-q'} \ln(1+(1-q')x)\right]^{1/(1-q)}.
\label{STexp}
\ee
Note that either parameter $q$ or $q'$  or both  take value one  the above functions simplifies to Tsallis $q$-functions (logarithm and exponential),   so we can write
\be
\log^{ST}_{q,1} (x)= \log^{ST}_{1,q} (x)  =\log^{T}_q (x), \; \log^{ST}_{1,1}(x)= \ln (x) \\
 \exp^{ST}_{q,1} (x)= \exp^{ST}_{1,q} (x)  =\exp^{T}_q (x), \;\; \exp^{ST}_{1,1} (x)= \exp(x).
 \label{STprop1}
\ee

\begin{figure}[htb]
	\begin{center}
		\includegraphics[width=.48\linewidth]{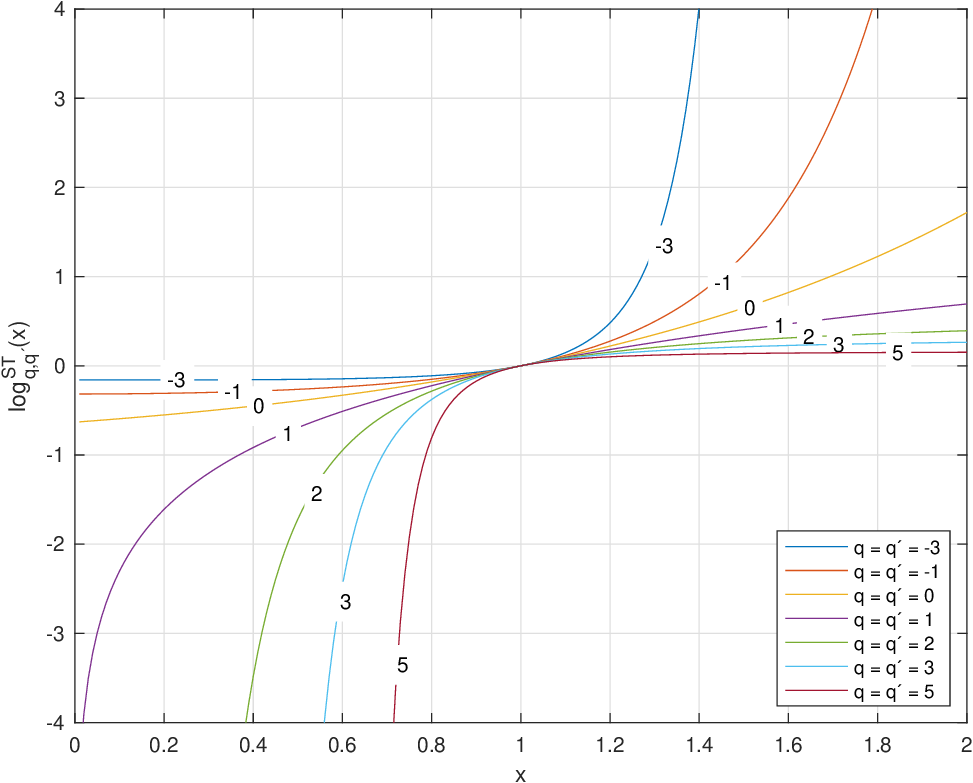}
		\hfill
		\includegraphics[width=.48\linewidth]{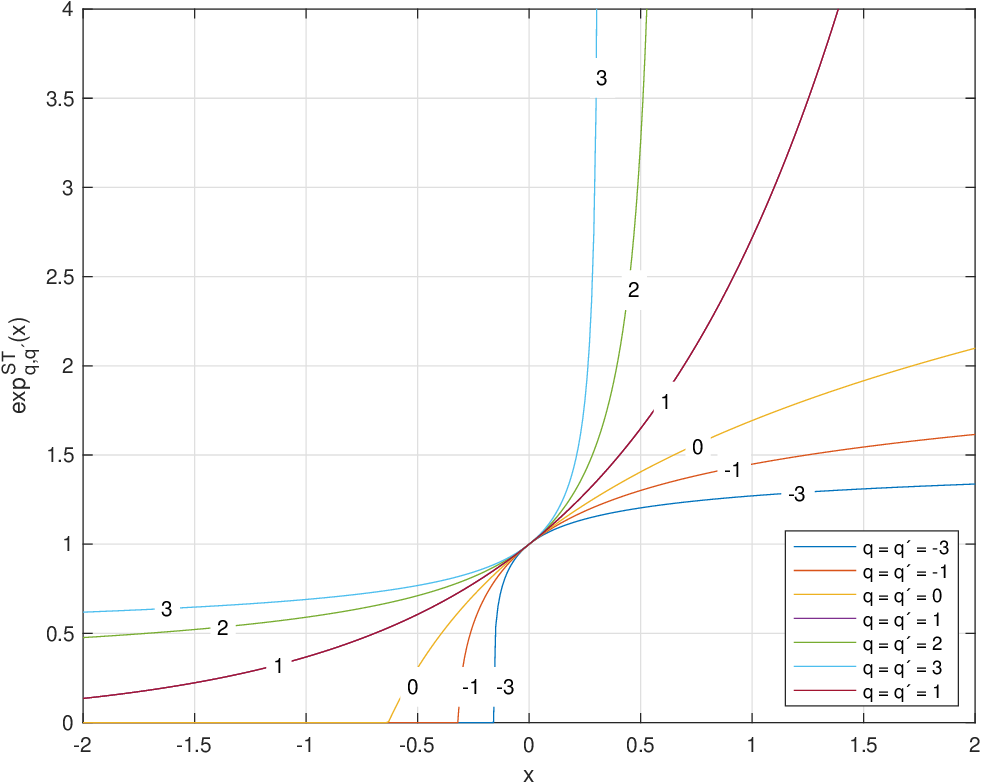}
		\caption{\color{black}Plots of the $(q,q')$-logarithm and $(q,q')$-exponential functions for different values of the parameters when $q=q'$.
		}
		\label{Fig-qq}
	\end{center}
\end{figure}

In the special case for $q=q'$, we obtain 
\be 
\log^{ST}_{q,q} (x)&=& \frac{1}{1-q} \left[ \exp (x^{1-q} -1) -1 \right] \nonumber \\
&=& \frac{1}{1-q} \left[ \exp \left((1-q) \log^T_q (x)\right)  -1\right], \quad x >0, \;\; q>0
\ee
and 
\be 
\exp^{ST}_{q,q} (x) = \left[\ln \left(x(1-q)+1\right)+1\right]_+^{1/(1-q)}.
\ee
 %
The plots of the $(q,q')$-logarithm and $(q,q')$-exponential for  various values of $q=q'$  are illustrated in Figure \ref{Fig-qq}.\\
%
%
Moreover, it easy to proof the following useful properties
\be
\log_{q,q'}^{ST} (1/x) = - \log^{ST}_{2-q,2-q'} (x),
\label{STprop2}
\ee
\be
\frac{d \log_{q.q'}^{ST} (x)}{d x} = x^{-q} \exp \left(\frac{1-q'}{1-q} (x^{1-q}-1)\right),
\label{STprop3}
\ee
Defining $(q,q')$-product as \cite{cardoso_borges2008}:
\be
x \otimes_{q,q'}^{ST} y = \exp_{qq'} \left(\log_{q,q'} (x) + \log_{q,q'} (y)\right)
\label{STprod}
\ee
we  have the key formulas 
\be
\exp^{ST}_{q,q'}(x+y) &=& \exp^{ST}_{q,q'}(x) \;\; \otimes_{q,q'} \;\;  \exp^{ST}_{q,q'}(y),\\
\exp^{ST}_{q,q'}(\log_{q,q'}^{ST} (x) +y)&=& x \;\;\otimes_{q,q'}  \;\;  \exp^{ST}_{q,q'}(y).
\label{STexpx+y}
\ee
Let consider now that link function MD is defined componentwise as
\be
f_{q,q'}(\bw) = \log_{q,q'}^{ST} (\bw).
\ee
The  novel $(q,q')$--GEG update can take formally the following form
\begin{empheq}[box=\fbox]{align}
\bw_{t+1} = \bw_t \otimes_{q,q'}  \exp_{q,q'}^{ST} \left(-\eta \nabla_{\bi w} L(\bw_t) \right),
\end{empheq}
In this case update is more complex  than in the  previous case.
Alternative approach is to apply the MMD/NG formula (\ref{diagMD}):
\begin{empheq}[box=\fbox]{align}
\displaystyle	\bw_{t+1} = \left[\bw_t  -\eta \diag \left\{ \bw^q \odot \exp \left( \frac{1-q'}{1-q} (1-\bw^{1-q}\right)\right\}  \nabla_{\bi w} L(\bw_t)\right]_+  
\end{empheq}
which can be written  equivalently in a scalar form as
\begin{empheq}[box=\fbox]{align}	
w_{i,t+1} = \left[ w_{i,t} - \eta w_{i,t}^q \exp\left(\frac{1-q'}{1-q} (1-\bw_{i,t}^{1-q})\right) \frac{\partial L(\bw_t)}{\partial w_i}\right]_+ 
\end{empheq}

{\bf Remark: Extension to  Three Parameters $(q,q',r)$--Logarithm}.
Note that by using definition $[x]_q= \exp(\log_q^T (x))$ we can write the ST logarithm
in a compact form
\be
\log_{q,q'}^{ST}(x) = \frac{[x]_{q}^{1-q'} -1}{1-q'} =
\frac{[\exp(\log^{T}_q(x)]^{1-q'}-1}{1-q'}
\ee
Analogously, we can define
\be
[x]_{q,q'}= \exp(\log_{q,q'}^{ST} (x)).
\ee
Hence, we can formulate three parameters logarithm proposed by
in as \cite{corcino2020}
\be
\log_{q,q',r}^{CC}(x) = \frac{([x]_{q,q'}^{1-q'})^{1-r} -1}{1-r} =
\frac{(\exp(\log^{ST}_{q,q'}(x))^{1-r}-1}{1-r}
\ee

\begin{figure}[thb]
	\begin{center}
		\includegraphics[width=.48\linewidth]{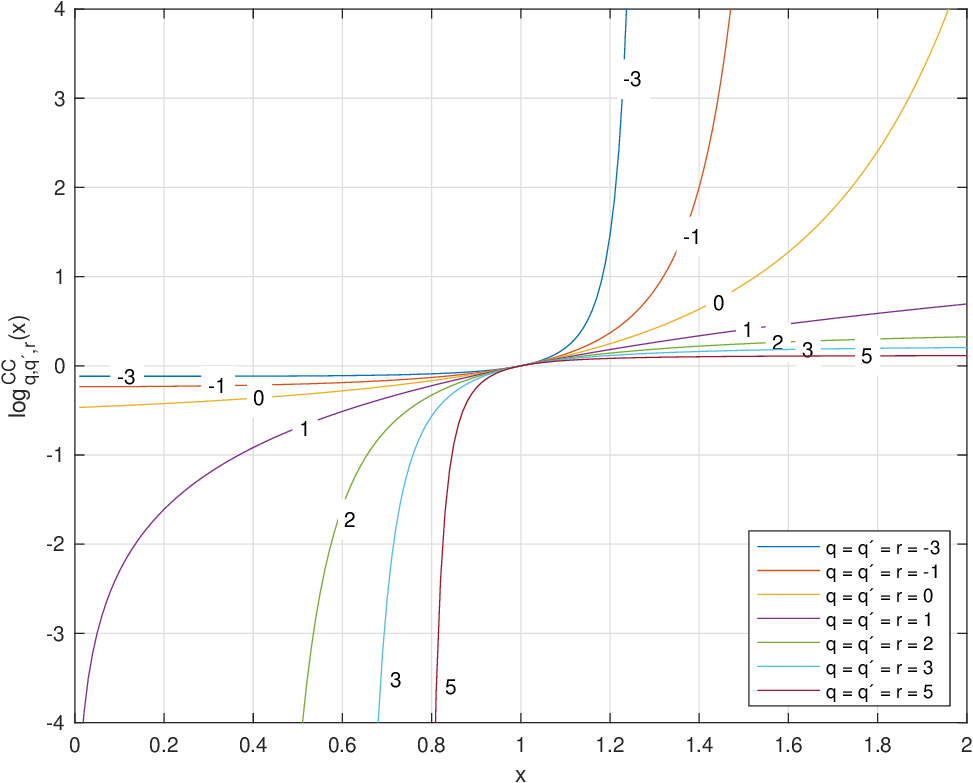}
		\hfill
		\includegraphics[width=.48\linewidth]{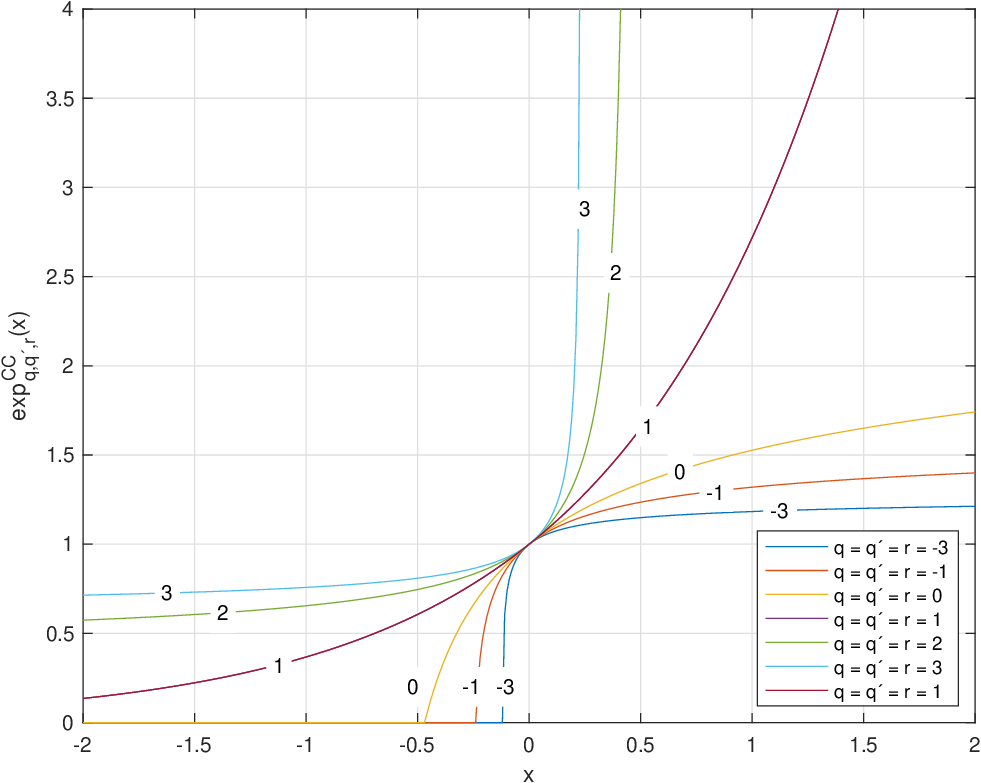}
		\caption{\color{black}Plots of the $(q,q',r)$-logarithm and $(q,q',r)$-exponential functions when the parameters are coincident $q=q'=r$.
		}
		\label{Fig_qqr}
	\end{center}
\end{figure}

The plots of the $(q,q',r)$-logarithm and $(q,q',r)$-exponential for  coincident values of $q$, $q'$, and $r$,  are illustrated in Figure \ref{Fig_qqr}.

In similar way, as in the previous section, we can drive MD updates using as a link function the above defined three parameter logarithm.

\section{{\bf MD and GEG  Using the Kaniadakis Entropy and its Extensions and Generalizations}}

\subsection{{\bf Basic Properties of $\kappa$-Logarithm and $\kappa$-Exponential}}

An entropic structure emerging in the context of special relativity, is the one defined by Kaniadakis \cite{kaniadakis_scarfone2002,kaniadakis2002} as follows
\be
S_{\kappa} (p_i) = \sum_i p_i \log^K_{\kappa}(1/p_i),
\ee
where deformed $\kappa$-logarithm referred to  as the Kaniadakis $\kappa$-logarithm is defined as \cite{kaniadakis_scarfone2002,kaniadakis2002}:
\be
 \label{deflogk}
	\log^K_{\kappa}(x)=\left\{
	\begin{array}{cl}
		\displaystyle \frac{x^{\kappa}- x^{-\kappa}}{2 \kappa} = \frac{1}{\kappa} \sinh (\kappa \ln (x)) & \text{if} \;\;  x>0 \;\;
\text{and} \;\;  0  <\kappa^2 <1\\
\\
		\ln (x) & \text{if} \;\; x>0 \;\; \text{and} \;\; \kappa=0.
	\end{array}
	\right.
\ee
The inverse function of the Kaniadakis $\kappa$-logarithm is the deformed exponential function $\exp^K_{\kappa} (x)$, represented as 
\be
 \label{defexpk}
	\exp^K_{\kappa}(x)&=&\exp \left( \int_{0}^{x}\frac{d y}{\sqrt{1+\kappa^2 y^2}}\right)\nonumber \\
&=&\left\{
	\begin{array}{cl}
		\displaystyle \left(\sqrt{1+\kappa^2 x^2} + \kappa x\right)^{1/\kappa}
= \exp\left(\frac{1}{\kappa} \; \text{arsinh} (\kappa x) \right)& -1 < \kappa < 1 \\
\\
		\exp(x) & \kappa =0.
	\end{array}
	\right.
\ee

\begin{figure}[htb]
	\begin{center}
		\includegraphics[width=.48\linewidth]{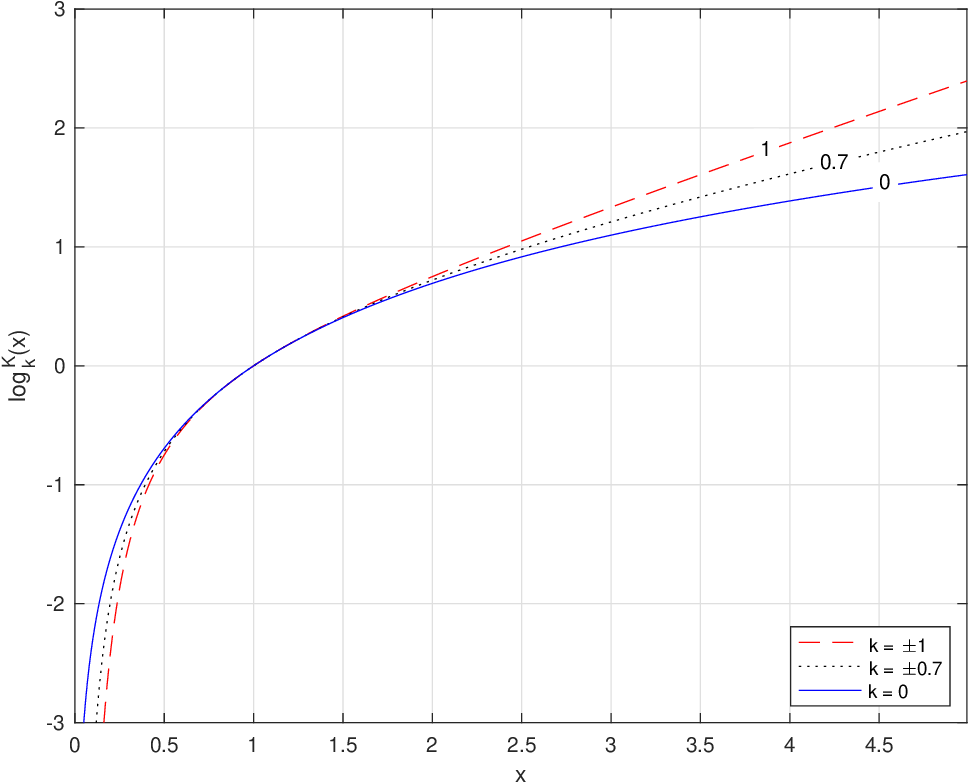}
		\hfill
		\includegraphics[width=.48\linewidth]{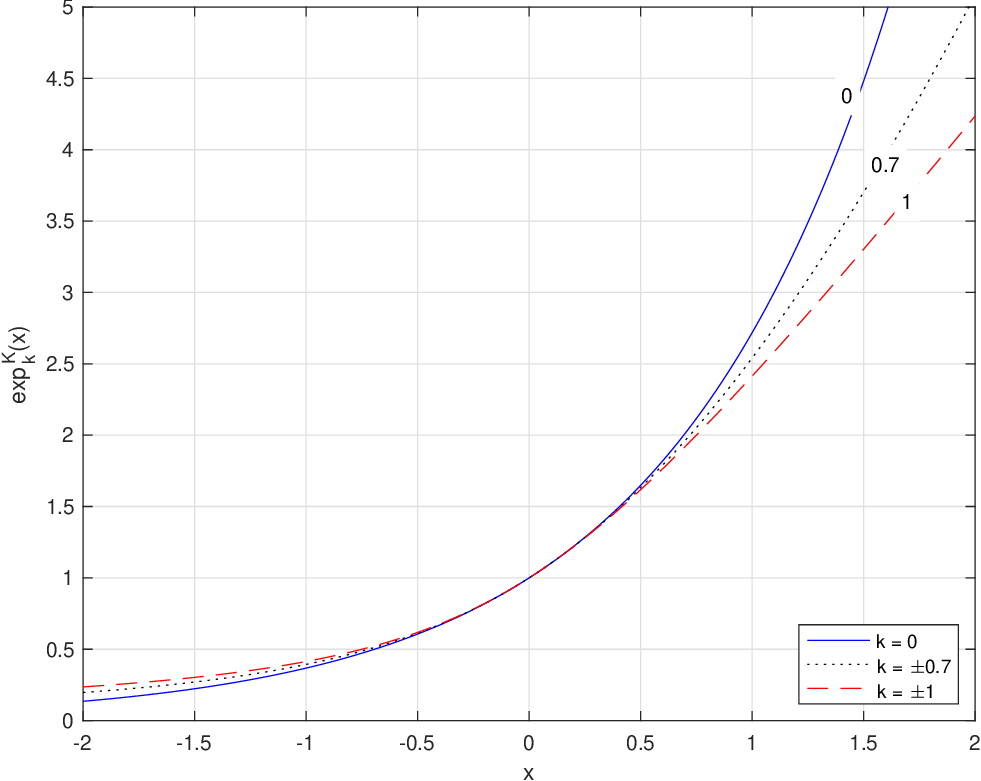}
		\caption{\color{black}Plots of the $\kappa$-logarithm and $\kappa$-exponential functions for different values of the parameter $\kappa$.}
		\label{Fig-kexp}
	\end{center}
\end{figure}

The plots of the $\kappa$-logarithm and $\kappa$-exponential functions for various values of $\kappa$ are illustrated in Figure \ref{Fig-kexp}.

Note that the Kaniadakis logarithm can be also expressed in terms of the Tsallis logarithm as 
\be
\log^K_{\kappa} (x) = \frac{\log^T_{1+\kappa} (x) + \log^T_{1-\kappa} (x)}{2}.
\ee
These functions have the following fundamental and useful properties 
\cite{kaniadakis2002,wada2023}:
\be
\log^K_{\kappa} (1) &=& 1, \; \log^K_{\kappa} (0^+) = -\infty, \; \log^K_{\kappa} (\infty) 
= +\infty,\\
\log^K_{\kappa} (1/x) &=& - \log^K_{\kappa} (x ), \\
\log^K_{\kappa} (x^{\lambda}) &=& \lambda \; \log^K_{\lambda \;\kappa} (x ), \\
\log^K_{\kappa} (x \; y) &=& y^{\kappa}\log^K_{\kappa} (x) + x^{-\kappa} \log^K_{\kappa}(y),\\
\log_{\kappa}(\exp(x)) &=& (1/\kappa) \sinh (\kappa x), \\
\ln(\exp_{\kappa}(x)) &=& (1/\kappa) \text{arsinh} (\kappa x),\\
\frac{\partial \log^K_{\kappa} (x)}{\partial x}  &=& \frac{x^{\kappa} + x^{-\kappa}}{2x} = \frac{\cosh[\kappa \ln (x)]}{x} > 0, \\
\frac{\partial^2 \log^K_{\kappa} (x)}{\partial x^2}  &=& \frac{\kappa-1}{2} x^{\kappa-2} - \frac{\kappa+1}{2} x^{-\kappa-2} <0  \;\;\text{for}  \;\;\kappa \in [-1,1].
\ee
The last two properties indicate that the Kaniadakis $\kappa$-logarithm is monotonically increasing and concave function.\\

The $\kappa$-logarithm can be approximated  as power series
\be
\log_{\kappa}(x) \approx \ln(x) + \frac{1}{ 3!} \kappa^2 [\ln(x)]^3 + \frac{1}{ 5!} \kappa^4 [\ln(x)]^5 + \frac{1}{ 7!} \kappa^6 [\ln(x)]^7 \cdots, \;\; x>0.
\ee
Furthermore, it is important to note that  applying the Taylor series expansion of the $\kappa$-exponential  we can obtain a simple approximation as
\be
\exp^K_{\kappa} (x) &=& 1+x + \frac{1}{2!} x^2 + \frac{1}{3!} (1-\kappa^2) x^3 + \frac{1}{4!} (1- 4 \kappa^4) x^4 +\cdots \\
&=& \exp(x) - \frac{1}{3!} \kappa^2 x^3 - \frac{1}{4!} 4 \kappa^4 x^4 +\cdots.
\ee
Two notable features of the $\kappa$-exponential function are that it asymptotically approaches a regular exponential function for small $x$ and asymptotically approaches a power law for large absolute $x$ \cite{kaniadakis2002,kaniadakis2009}. Specifically,
\be
\lim_{x \rightarrow 0} \exp_{\kappa} (x)  \sim \; \exp(x), \\
\lim_{x \rightarrow \pm \infty} \exp_{\kappa} (x)  \sim  \; |2 \kappa x|^{\pm 1/|\kappa|}.
\ee
The $\kappa$-exponential function has the following basic properties \cite{kaniadakis_scarfone2002,kaniadakis2002,wada2023}
\be
\exp^K_{\kappa}(0) &=& 1, \;\; \exp^K_{\kappa}(-\infty)= 0^+,\;\; \exp^K_{\kappa}(\infty)=+\infty, \\
 \exp^K_{\kappa}(-\infty) &=& 0^+, \;\; \exp^K_{\kappa}(+\infty)= +\infty,\\
  \exp^K_{\kappa}(x)\exp^K_{\kappa}(-x) &=& 1, \\
\left[\exp^K_{\kappa}(x)\right]^r &=& \exp^K_{\kappa/r}(rx), \;\; r \in \Real,\\
  \frac{\partial \exp^K_{\kappa} (x)}{\partial x} & >& 0,\\
\frac{\partial^2 \exp^K_{\kappa} (x)}{\partial x^2} & >0&  \;\;\text{for} \;\;\kappa \in [-1,1].
  \label{expkprop}
\ee
The last two properties means that the Kaniadakis $\kappa$-exponential is monotonically increasing and convex function.

The property (\ref{expkprop}) emerges as particular case of the more general one
\be
\exp^K_{\kappa}(x)\exp^K_{\kappa}(y) &=& \exp^K_{\kappa}(x) \oplus_{\kappa} \exp^K_{\kappa}(y),\\
\log^K_{\kappa}(x y )&=& \log^K_{\kappa}(x) \oplus_{\kappa} \log^K_{\kappa}(y),
\ee
where $\kappa$-addition is defined as
\be
x \oplus_{\kappa} y &= & x \; \sqrt{1+\kappa^2 y^2}   +  y \; \sqrt{1+\kappa^2 x^2} \\
&\approx& x+y + \frac{\kappa^2}{2} (x y^2 +x^2 y) - \frac{\kappa^4}{8} (x y^4 + x^4 y) \cdots
\ee
By defining and evaluating the $\kappa$-product
\be
\displaystyle x \otimes_{\kappa} y &= &\exp_{\kappa} \left[ \log_{\kappa} (x) + \log_{\kappa} (y)\right] \\
&=& \left(\frac{x^{\kappa} -x^{-\kappa}}{2} + \frac{y^{\kappa} -y^{-\kappa}}{2} +  \; \sqrt{1+ \left(\frac{x^{\kappa} -x^{-\kappa}}{2} + \frac{y^{\kappa} -y^{-\kappa}}{2}\right)^2 }\right)^{1/\kappa} \\
&=& \exp\left(\frac{1}{\kappa}\text{arsinh}((x^{\kappa} - x^{-\kappa} + y^{\kappa} - y^{-\kappa})/2) \right) \\
&=& \frac{1}{\kappa} \sinh \left( \frac{1}{\kappa} \text{arsinh}(\kappa x) \; \text{arsinh}(\kappa y)\right),
\label{kprod}
\ee
we  have the key formulas for our MD application
\be
\exp^K_{\kappa}(x+y) = \exp^K_{\kappa}(x) \otimes_{\kappa} \exp^K_{\kappa}(y),\\
\exp^K_{\kappa}(\log^K_{\kappa} (x)+y) = x \otimes_{\kappa} \exp^K_{\kappa}(y).
\ee

Appendix~\ref{sec:kappaalgebra} gives an overview of the $\kappa$-algebra and calculus.

\subsection{{\bf MD  and GEG Using the Kaniadakis Entropy and $\kappa$-Logarithm}}

Let assume that the link function in Mirror Descent can take the following componentwise form
\be
f_{\kappa}(\bw_t) = \log^K_{\kappa} (\bw_t), \qquad \bw=[w_{1,t}, \ldots, w_{N,t}]^T \in \Real_+^N.
\label{linkfk}
\ee
Note that since the first derivative of $\log^K_{\kappa} (\bw)$ is positive and second derivative is negative the link function is (componentwise) increasing  and concave function for $\kappa \in (0,1]$.
In this case the  mirror map:
\be
F_{\kappa}(\bw_t)= \sum_{i=1}^N \frac{1}{2 \kappa} \left[ \frac{w_{i,t}^{1+\kappa}}{1+\kappa} -  \frac{w_{i,t}^{1-\kappa}}{1-\kappa} \right].
\ee

Taking into account formula (\ref{f-1fDT}),  we obtain novel $\kappa$-GEG  update
\begin{empheq}[box=\fbox]{align}
\bw_{t+1} = \bw_t \; \otimes_{\kappa} \; \exp_{\kappa}^K \left(-\eta \nabla_{\bi w} L(\bw_t) \right)
\label{kEG1}
\end{empheq}

where $ \otimes_{\kappa}$ is $\kappa$-product  defined by Eq. (\ref{kprod}) is performed  componentwise.

\subsection{\bf {Two-Parameters  Logarithms  Based on  Generalized  Kaniadakis-Lissia-Scarfone Entropy}}

Another important  generalized entropy has the following form
\be
S_{\kappa, r} (p) = - \sum_{i=1}^W (p_i)^{r+1} \frac{p_i^{\kappa}- p_i^{-\kappa}}{2 \kappa}
= - \sum_{i=1}^W p_i \log_{\kappa,r} (p_i),
\label{entropyKLS}
\ee
which was introduced by Sharma, Taneja and Mittal (STM) in \cite{sharma1975,mittal1975}, and also investigated, independently,  by Kaniadakis, Lissia and Scarfone (KLS)
in \cite{kaniadakis2004,kaniadakis2005}. 

Equation (\ref{entropyKLS}) mimics the expression of the Boltzmann-Gibbs entropy by replacing the standard natural logarithm $\ln (x) $ with the two-parametric deformed logarithm $\log_{\kappa,r} (x)$ defined as
\be
\log_{\kappa,r} (x) = x^r  \; \frac{x^{\kappa}- x^{-\kappa}}{2 \kappa}, \;\; x>0, \;\; r  \in \Real,\; \; \;\;\text{for}  \;\;\
-|\kappa| \leq r \leq \1/2 - |1/2- |\kappa| |.
\ee

\begin{figure}[htb]
	\begin{center}
		\includegraphics[width=.48\linewidth]{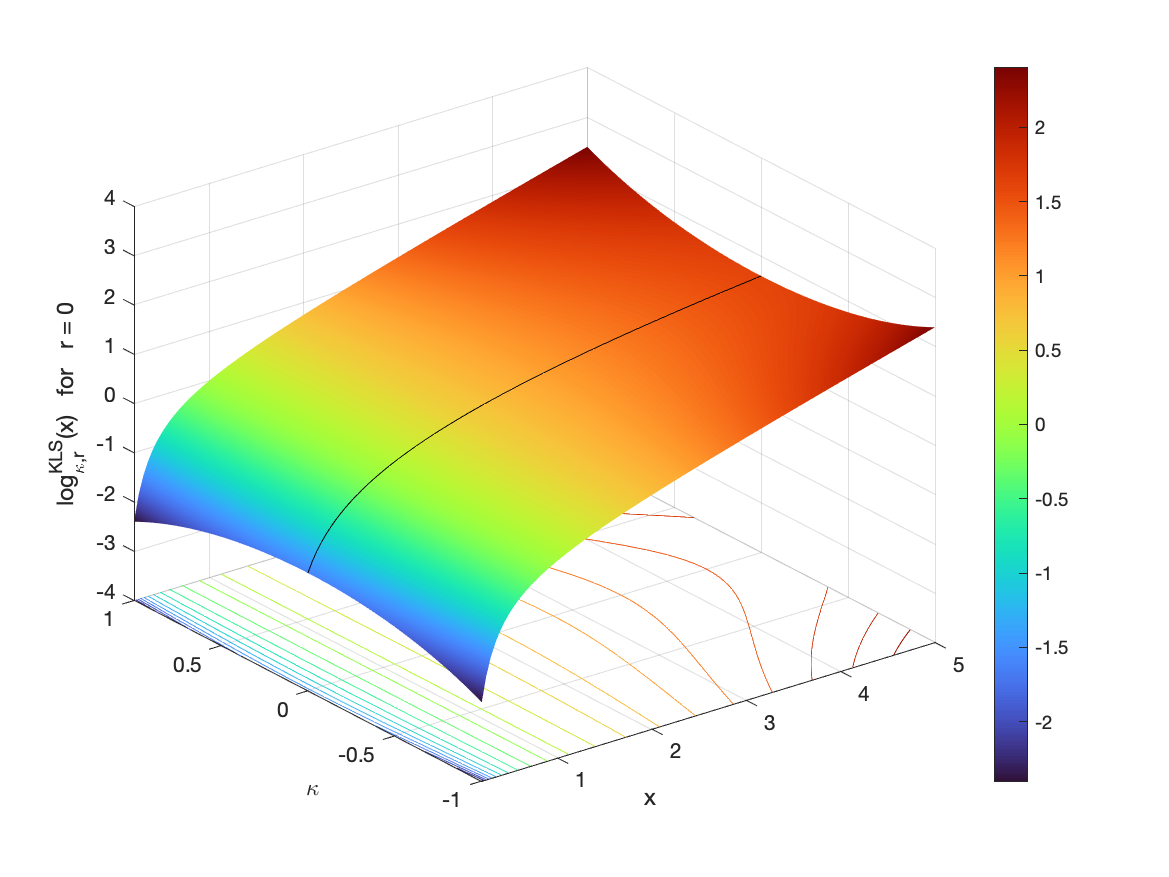}
		\hfill
		\includegraphics[width=.48\linewidth]{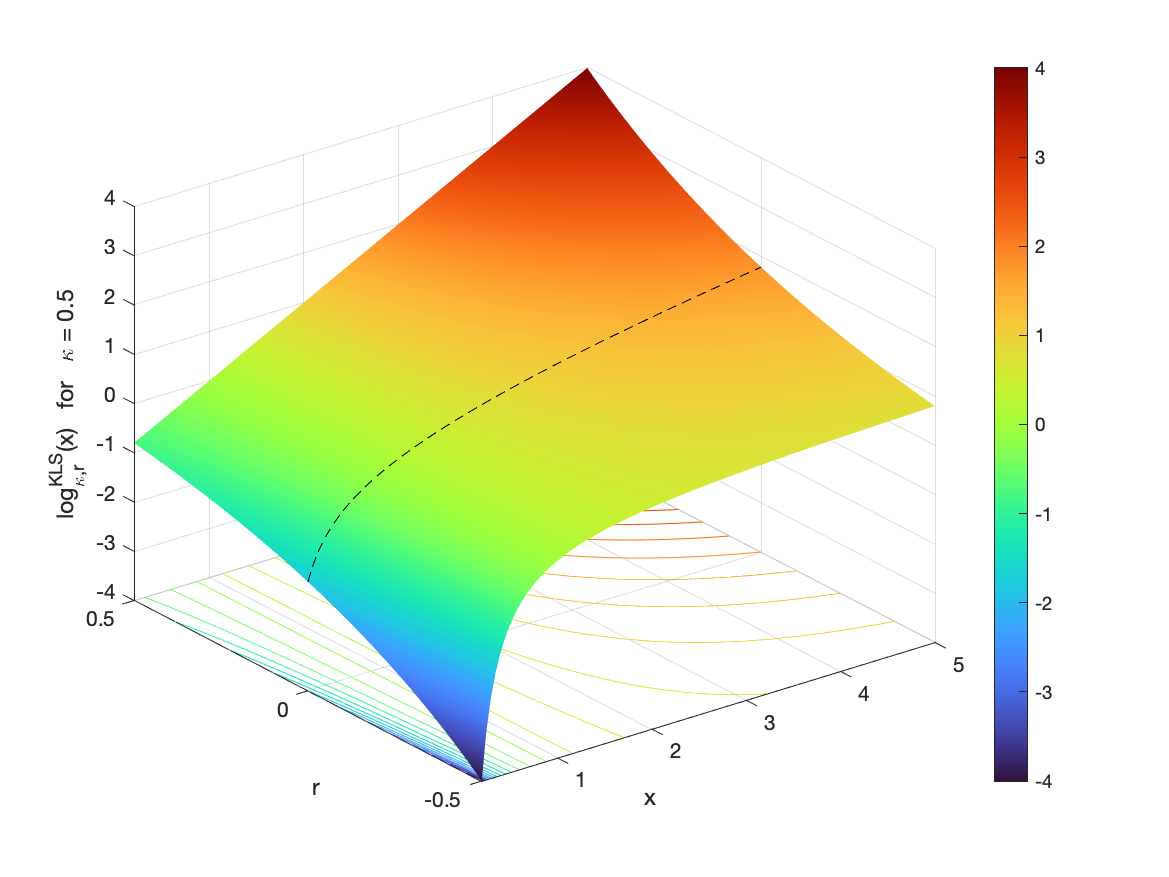}
		\caption{\color{black} Surface plots of the $(\kappa,r)$-logarithm for  various values of hyperparameters $\kappa$ and $r$. The  left-hand-side figure illustrate the $(\kappa,r)$-logarithm in terms of $\kappa$ and $x$ when $r=0$, therefore, it coincides with the $\kappa$-logarithm. The black continuous line represents the reference of the classical logarithm, which is obtained for $\kappa=0$. The  right-hand-side figure illustrates the  $(\kappa,r)$-logarithm, now in terms of $r$ and $x$, when $\kappa=0.5$. The black dashed line represents the $\kappa$-logarithm for $\kappa=0.5$.}
	
	\label{Fig-krlog-a}
\end{center}
\end{figure}

\begin{figure}[htb]
	\begin{center}
		\includegraphics[width=.48\linewidth]{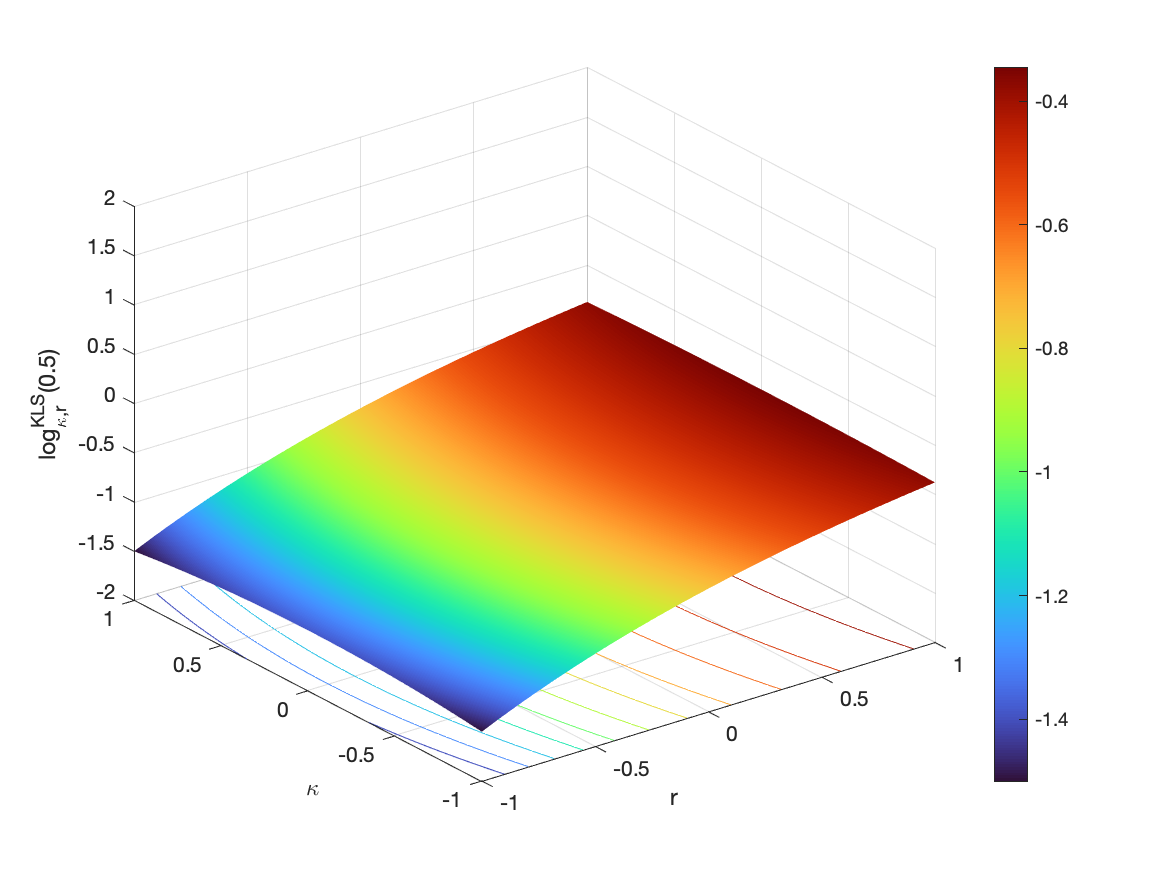}
		\hfill
		\includegraphics[width=.48\linewidth]{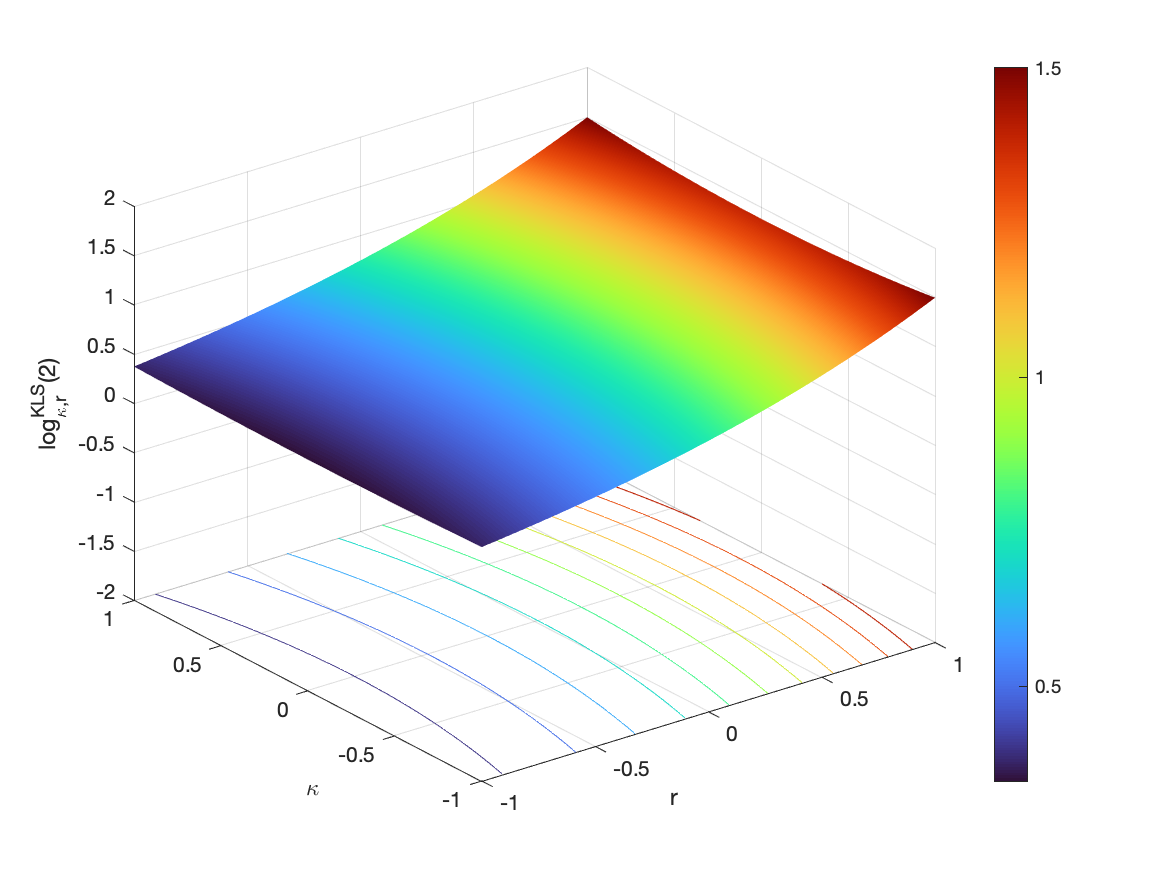}
		\caption{\color{black} Surface plots of the $(\kappa,r)$-logarithm in terms of  the hyperparameters $\kappa$ and $r$, when $x$ is fixed. The figure in the left-hand-side evaluates the $(\kappa,r)$-logarithm for $x=0.5$, whereas the figure in the right-hand-side evaluates it for $x=2$. From the drawings, it is apparent how the changes of the hyperparameter $r$ have a much stronger influence in the magnification of the response, in comparison with the changes of the hyperparameter $\kappa$ that correspond with more subtle elongations.}	
		\label{Fig-krlog-b}
	\end{center}
\end{figure}

The surface plots of the $(\kappa,r)$-logarithm for various values of hyperparameters $\kappa$ and $r$ are illustrated in Figures \ref{Fig-krlog-a}-\ref{Fig-krlog-b}.

\color{black}

The $(\kappa,r)$-logarithm can be expressed via the Kaniadakis $\kappa$-logarithm and the Tsallis $q$-logarithm
\be
\log_{\kappa,r} (x) = x^r \; \frac{x^{\kappa}- x^{-\kappa}}{2 \kappa} = x^r \; \log^K_{\kappa} (x) = x^{r-\kappa} \log^T_{1- 2 \kappa} (x).
\ee
Obviously, for $r=0$ the $(\kappa,r)$-logarithm simplifies to Kaniadakis $\kappa$-logarithm, and for $r= \pm |\kappa|$ one recovers the Tsallis $q$-logarithm with $q= 1 \mp 2 |\kappa|$, ($0<q<2$).

By introducing a new parameter $\omega= r/\kappa$ or replacing $r=\omega \; \kappa$  we
can represent the logarithm as
\be
\log_{\kappa,\omega} (x) =  \frac{x^{\kappa(\omega+1)}- x^{\kappa(\omega-1)}}{2 \kappa},
\ee
which for $\omega=0$ simplifies to $\kappa$-logarithm and $\omega =1$ and $\kappa= (1-q)/2$ we have $q$-logarithm. This formula indicates that the this logarithm smoothly interpolates between Kaniadakis logarithm and Tsallis logarithm.\\

Summarizing, the $(\kappa,r)$-logarithm can be described as follows \cite{kaniadakis2004,kaniadakis2005}
\be
 \label{deflogKLS}
	\log^{KLS}_{\kappa,r}(x)=\left\{
	\begin{array}{cl}
		\displaystyle \frac{x^{r+\kappa}- x^{r-\kappa}}{2 \kappa} & \text{if} \;\; x>0 \;\;
\text{for} \;\; r  \in \Real,\; \;\text{and}  \;\;\
 -|\kappa| \leq r \leq |\kappa|,\\
\\
\displaystyle \log^K_{\kappa} (x) = \frac{x^{\kappa} - x^{-\kappa}}{2 \kappa} & \text{if} \;\; x>0 \;\;
\text{for} \; r=0, \; \kappa \in [-1,1], \;\; \kappa \neq 0, \\
\\
\displaystyle \log^T_q (x) = \frac{x^{1-q} -1}{1-q}, & \text{if} \;\; x>0 \;\;
\text{for} \; r= \kappa=(1-q)/2, \;\; q \neq 1, \; q>0,\\
\\
		\ln (x) & \text{if} \;\; x>0 \;\; \text{and} \;\; r=\kappa=0.
	\end{array}
	\right.
\ee
It  should be also noted the  $(\kappa,r)$-logarithm can be represented approximately by the following power series
\be
\displaystyle \log^{KLS}_{\kappa,r} \approx \ln (x) + \frac{1}{2} (2 r) [\ln (x)]^2 + \frac{1}{6}  (\kappa^2 + 3 r^2 ) [\ln(x)]^3 \cdots.
\ee
The $(\kappa,r)$ logarithm has the following basic properties
\be
\log_{\kappa,r} (1) &=& 0, \; \log_{\kappa,r} (0^+) = -\infty,\; 
 \text{for}\;r<|\kappa|, \; \log_{\kappa,r} (+\infty) = +\infty, \; \text{for} \; r> -|\kappa|,
 \nonumber \\
\log_{\kappa,r} (x) &=& - \log_{\kappa,-r} (1/x )= \log_{-\kappa,r} (x ), \\
\log_{\kappa,r} (x^{\lambda}) &=& \lambda \; \log_{\lambda \;\kappa} (x ), \\
\frac{\partial \log_{\kappa,r} (x)}{\partial x}  &>0&,  \;\;\text{for} \;\;
 -|\kappa| \leq r \leq |\kappa|,\\
\frac{\partial^2 \log_{\kappa,r} (x)}{\partial x^2}  &<0&  \;\;\text{for}  \;\;\
-|\kappa| \leq r \leq \1/2 - |1/2- |\kappa| |.
\ee
The last two properties indicate that the  $(\kappa,r)$-logarithm is monotonically increasing and concave function.\\
{\bf Remarks}:

{\bf Relation to the Euler Logarithm} It interesting to note  that for $r+k =a$ and $r-k=b$ the KLS $\kappa,r$--logarithm can be represented as the Euler logarithm \cite{CichockiEU}:
\be 
\log_{a,b}^{Eu} (x) = \frac{x^a-x^b}{a-b}, \qquad x >0, \;\;  a<0,  \;\;  0 <b < 1, 
\label{Eulog}
\ee
which is related to the Borges--Rodity entropy \cite{Borges1998,furuichi2010}.\\

{\bf Connection to the Schw\"ammle-Tsallis logarithm}: By applying nonlinear transformation in (\ref{Eulog})  $x \rightarrow \exp(\log_q(x))$ and $a=1-q'$, $b=0$, we obtain the Schw\"ammle--Tsallis logarithm.\\

{\bf Connection to the Mean Value Theorem}:
The function has deep connections to the Mean Value Theorem applied to power functions.
 For the power function $ g(t) = x^t $, the Mean Value Theorem guarantees the existence of some parameter $ c \in (a,b) $ such that
$ g'(c) = \frac{g(b) - g(a)}{b - a} $,
this yields: 
\be 
\log^{EU}_{a,b}(x) = \frac{x^b - x^a}{b - a} = c \cdot x^{c-1} \cdot \ln(x) \;\;\text{for} \;\; c \in (a,b). 
 \ee
{\bf Logarithmic Mean Connection}:
The function relates to the logarithmic mean $ L(u,v) = \frac{u-v}{\ln u - \ln v} $ through the substitution $ u = x^a, v = x^b $. 
These connections provide  alternative computational approaches and theoretical insights.\\
{\bf Exponential Function Theory}:
The underlying structure connects to exponential function differentiation rules, where $ \frac{d}{dt}x^t = x^t \ln(x) $, 
explaining the limiting behavior observed in the analysis.\\
{\bf  Computational and numerical considerations}: 
The numerical analysis reveals several important computational aspects:
\begin{enumerate}
  \item 	Numerical Stability: The KLS logarithm becomes increasingly stable as $ x \to 1 $, 
  but exhibits potential numerical instability for $ x $ values far from unity.
  
  \item Parameter Sensitivity: Small $ x $ values create higher sensitivity to parameter changes, 
  requiring careful numerical handling.
  
  \item Convergence Properties: The limiting behavior  requires special computational treatment using L'Hôpital's rule.
\end{enumerate}

\subsection{\bf Exponential KLS Function and its Properties}

The existence of $\exp_{\kappa,r}(x)$, the inverse function of $\log_{\kappa,r}(x)$ follows from its monotonicity in $\Real$ although an explicit expression, in general, cannot be given.
In other words, the inverse function can not be expressed in  a closed analytical form but it can be approximated and expressed for example  in terms of the Lambert-Tsallis $W_q$-functions, which are the solution of equations $W_q(z)[ 1+(1-q) W_q(z)]_+^{1/(1-q)} = z$ :
\be
\displaystyle \exp_{\kappa,r} (x) = \left( \frac{W_{\frac{\lambda+1}{\lambda}} \left(\lambda \tilde{x}^{-\lambda} \right)}{\lambda}\right)^{-1/(2\kappa),}
\ee
where $\lambda = (2\kappa)/(r+\kappa)$, $\tilde x = 2 \kappa x$ and $W_q$ is the Lambert-Tsallis function \cite{lambert_tsallis}.\\
Another much simpler approach is to use Lagrange's inversion theorem around 1
 to obtain the following rough power series approximation (which may be  sufficient for  the most our applications):
\be
\displaystyle \exp_{\kappa,r} (x) &\approx& 1+x + \frac{1}{2} (1-2r) x^2 + \left(\frac{1}{6}-r + \frac{3}{2}r^2 - \frac{1}{6}\kappa^2\right) x^3  + \cdots\\
&=& \exp (x) - rx^2 +  \left(\frac{3}{2} r^2 - r - \frac{1}{6}\kappa^2\right) x^3 + O(x^4).
\ee
Hence,  we can represent $(\kappa,r)$-exponential as follows:
\be
 \label{defexpKLS}
	\exp_{\kappa,r}(x)=\left\{
	\begin{array}{cl}
		\approx \displaystyle \exp(x) - r x^2  +  \left(\frac{3}{2} r^2 - r - \frac{1}{6}\kappa^2 \right) x^3   & \text{for} \; r  \in \Real,\; \;
 -|\kappa| \leq r \leq |\kappa|, \;\;|\kappa| <1,\\
\\
\displaystyle \exp^K_{\kappa} (x) = \left[\kappa x + \sqrt{1+\kappa^2 x^2}\right]^{1/\kappa} & \;\;
\text{for} \; r=0, \; \kappa \in [-1,1], \;\;\kappa \neq 0, \\
\\
\displaystyle \exp^T_q (x) = \left[ 1+(1-q)x^{} \right]_+^{1/(1-q)} &
\text{for} \; r= \kappa=(1-q)/2, \;\; q \neq 1, \\
\\
		\exp (x) & \text{for} \;\; r=\kappa=0.
	\end{array}
	\right.
\ee

Furthermore, the $(\kappa,r)$-exponential function has the following fundamental properties:
\be
\exp_{\kappa,r}(0) &=& 1, \;\;\exp_{\kappa,r}(-\infty) = 0^+,\;\text{for} \; r < |\kappa|, \; \exp_{\kappa,r}(_\infty)=+\infty, \;\; \text{for} \;\; r > -|\kappa|, 
\nonumber \\
 \exp_{\kappa,r}(x)\exp_{\kappa,r}(-x) &=& 1,\\
(\exp_{\kappa,r}(x))^{\lambda} &=&
\exp_{\kappa/\lambda,r/\lambda}(\lambda x),\\
  \frac{\partial \exp_{\kappa,r} (x)}{\partial x} & >& 0, \;\;\text{for} \;\;
 -|\kappa| \leq r \leq |\kappa|,\\
\frac{\partial^2 \exp_{\kappa,r} (x)}{\partial x^2} &>& 0, \;\;\text{for}  \;\;\
-|\kappa| \leq r \leq 1/2 - | 1/2- |\kappa| |
  \label{expKLS}
\ee
The last two properties means that the  $(\kappa,r)$-exponential is monotonically increasing and convex function.\\
Two notable features of the $(\kappa,r)$-logarithm and exponential function are that it asymptotically approaches a regular exponential function for small $x$ and asymptotically approaches a power law for large absolute $x$:
\be
\lim_{x \rightarrow 0^+} \log_{\kappa,r} (x)  &\sim& \; \frac{-1}{2 |\kappa| \;x^{|\kappa|+r}}, \\
\lim_{x \rightarrow +\infty}\log_{\kappa,r} (x)  &\sim&  \; \frac{x^{|\kappa|+r}}{2\kappa},\\
\lim_{x \rightarrow 0} \exp_{\kappa,r} (x)  &\sim& \; \exp(x), \\
\lim_{x \rightarrow \pm \infty}\exp_{\kappa,r} (x)  &\sim&  \; |2 \kappa x|^{1/(r \pm |\kappa|)}.
\ee
By defining
\be
\displaystyle x \otimes_{\kappa,r} y = \exp_{\kappa,r} \left[ \log_{\kappa,r} (x) + \log_{\kappa,r} (y)\right]
\ee
we  have the key formulas for our MD  (GEG) implementations"
\be
\exp_{\kappa,r}(x+y) &=& \exp_{\kappa,r}(x) \otimes_{\kappa,r} \exp^K_{\kappa,r}(y),\\
\exp_{\kappa,r}(\log^K_{\kappa,r} (x)+y) &=& x \otimes_{\kappa,r} \exp^K_{\kappa,r}(y).
\ee
Let assume that the link function is defined as  $f(\bw)= \log_{\kappa,r}(\bw) $ and its inverse (if approximated version is accepted) $ f^{(-1)} (\bw) = \exp_{\kappa,r}(\bw)$, then
using a general MD formula (\ref{f-1fDT}), and fundamental properties described above, we obtain a general MD formula employing a wide family of deformed  logarithms arising from
group entropies or trace-form entropies:
%
\begin{empheq}[box=\fbox]{align}
\displaystyle \bw_{t+1} = \exp_{\kappa,r} \left[\log_{\kappa,r}(\bw_t) - \eta_t \nabla L(\bw_t)\right] =\bw_t  \; \otimes_{\kappa,r} \; \exp_{\kappa,r}\left(-\eta_t \nabla L(\bw_t)\right)
\end{empheq}
where $\otimes_{\kappa,r}$-multiplication is defined/determined as follows
\be
 x \; \otimes_{\kappa,r} \; y  = \exp_{\kappa,r} \left(\log_{\kappa,r}(x) \; + \; \log_{\kappa,r}(y)\right).
\ee
Alternatively, due to some complexity of computing precisely  $\exp_{\kappa,r} (x)$, in general case, we can use formula MMD/NG (\ref{diagMD}) to derive, a quite flexible and general NG gradient update:
\begin{empheq}[box=\fbox]{align}
\displaystyle \bw_{t+1} = \left[\bw_t  - \eta_t \diag \left\{ \left(\frac{\partial \log_{\kappa,r} (\bw_t)}{
\partial \bw_t}\right)^{-1} \right\} \left(\nabla L(\bw_t)\right) \right]_+
\end{empheq}
where  $\displaystyle \diag \left\{ \left(\frac{\partial \log_{\kappa,r} (\bw_t)}{ \partial\bw_t} \right)^{-1} \right\}$ is a  positive--definite diagonal matrix, with the diagonal entries
\be
\displaystyle \left(\frac{\partial \log_{\kappa,r} (\bw_{t})}{ \partial w_{i,t}} \right)^{-1} = \frac{2 \kappa}{(r+\kappa)  w_{i,t}^{r+\kappa-1} - (r-\kappa) w_{i,t}^{r-\kappa-1}} >0, \;  -|\kappa| \leq r \leq |\kappa|.
\ee

\section{Generalization and Normalization of Mirror Descent}

Summarizing, all of the GEG updates proposed  in this paper 
can be presented in normalized form (by projecting to unit simplex) 
in the following general and flexible form
\be
\tilde{\bw}_{t+1} &=& \bw_t \otimes_D \exp_D\left(-\bm\eta_t  \odot \nabla \widehat{L} (\bw_t) \right)\qquad \text{(Generalized multiplicative update)},\\
\bw_{t+1} &=& \frac{\tilde{\bw}_{t+1}}{||\tilde{\bw}_{t+1}||_1},   \qquad \text{(Projection to unit simplex)}
\ee
where  $\exp_D(x)$ ($\log_D(x)$) is a generalized exponential (logarithm),  $\widehat{L} (\bw_t) = L(\bw_t/||\bw_t||_1)$ is normalized/scaled loss function, $\bm\eta_t$ is a vector of the  learning rates,  $\nabla \widehat{L} (\bw_t) = \nabla {L} (\bw_t) - (\bw^T \nabla {L} (\bw_t)){\bf 1}\;\;$  and the generalized $D$-multiplication is computed as
\be
\bw_t \otimes_D \exp_D (\bg_t) = \exp_D(\log_D (\bw_t) + \bg_t).
\ee
Here, $\log_D(\bw_t)$ and its inverse $\exp_D(\bw_t)$ mean any deformed  logarithm  and exponential  investigated in this paper (i.e., the  Tsallis, Kaniadakis, ST,  KLS and KS exponential/logarithm).

Alternatively, when inverse function can not be precisely computed, we can use MMD/NG additive natural gradient formula (\ref{diagMD}), which is expressed in general for as 
\begin{empheq}[box=\fbox]{align}
\bw_{t+1} &= \left[\bw_t  -\eta \diag \left\{\left(
\frac{d\,\log_D(\bw_t)}{d \, \bw_t}\right)^{-1}\right\} \nabla_{\bi w} \widehat{L}(\bw_t)\right]_+,\\
\bw_{t+1} &= \frac{\tilde{\bw}_{t+1}}{||\tilde{\bw}_{t+1}||_1}, \quad \quad  \bw_t \in \Real_+^N,\; \forall t
		\label{diagMDL}
\end{empheq}
where $ \displaystyle \diag \left\{ \left(\frac{d\,\log_D(\bw)}{d \bw}\right)^{-1}\right\} = \diag \left\{ \left(\frac{d\,\log_D(\bw)}{d w_1} \right)^{-1}, \ldots, \left(\frac{d\,\log_D(\bw)}{d w_N}\right)^{-1} \right\} $ 
is a diagonal positive--definite matrix.

\begin{sidewaystable}

\begin{tabular}{|l|l|l|}\hline
Entropy  & Deformed exponential & MD/GEG update \\ \hline
Shannon &  $\exp(x)=\sum_{i=0}^\infty \frac{x^i}{i!}$ 
& $\bw_{t+1} = \bw_{t}\odot \exp(-\eta \nabla L(\bw_t))$ (\mbox{EG})\\ \hline
Tsallis & 
$\exp^T_{q}(x)=\left\{
	\begin{array}{cl}
		  [1+ (1-q) x]_+^{1/(1-q)} & q \neq 1\\
 \\ \hline 
		\exp(x) & q=1
	\end{array}
	\right.$  &
	$\bw_{t+1} =  \exp_q^T \left[ \log_q^T (\bw_t) \otimes_q   \left(-\eta \nabla_{\bi w} L(\bw_t) \right) \right]$ \\
	&& $=  \bw_t \otimes_q  \exp_q^T \left(-\eta \nabla_{\bi w} L(\bw_t) \right)$   (\mbox{$q$-GEG})\\ \hline
Schw{\"a}mmle-Tsallis &
$\exp_{q,q'}^{\ST} (x) = \left[ 1+ \frac{1-q}{1-q'} \ln(1+(1-q')x)\right]^{1/(1-q)}$ 
&
$\bw_{t+1} = \left[\bw_t  -\eta \diag \left\{ \bw^q \odot \exp \left( \frac{1-q'}{1-q} (1-\bw^{1-q}\right)\right\}  \nabla_{\bi w} L(\bw_t)\right]_+$
\\ \hline
Kanadiakis &
$\exp^K_{\kappa}(x)=\left\{
	\begin{array}{cl}
		\text{arcsinh} (\kappa x)  & -1 < \kappa < 1 \\
		\exp(x) & \kappa =0.
	\end{array}
	\right.$
&
$\bw_{t+1} = \bw_t \; \otimes_{\kappa} \; \exp_{\kappa}^K \left(-\eta \nabla_{\bi w} L(\bw_t) \right) (\mbox{$\kappa$-GEG})$
\\ \hline
KLS & 
$\exp_{\kappa,r} (x)$
&
$ \bw_{t+1} = \left[\bw_t  - \eta_t \diag \left\{ \left(\frac{\partial \log_{\kappa,r} (\bw_t)}{
\partial \bw_t}\right)^{-1} \right\} \left(\nabla L(\bw_t)\right) \right]_+$\\ \hline
Generic &
$\exp_D(x)$
& 
$\bw_{t+1} = \left[\bw_t  -\eta \diag \left\{\left(
\frac{d\,\log_D(\bw_t)}{d \, \bw_t}\right)^{-1}\right\} \nabla_{\bi w} \widehat{L}(\bw_t)\right]_+$\\
& &
$\bw_{t+1} = \frac{\tilde{\bw}_{t+1}}{||\tilde{\bw}_{t+1}||_1}$\\ \hline
\end{tabular}

\caption{Overview of the generalized exponentiated gradient (GEG) updates.\label{tab:GEG}}

\end{sidewaystable}

\section{Conclusion and Discussions}

This study establishes a comprehensive framework for applying trace-form entropies and associated deformed logarithms in both Mirror Descent and  equivalently Generalized Exponentiated Gradient algorithms. By systematically exploring trace--form entropies, especially Tsallis, Kaniadakis, Scarfone, and Sharma-Taneja-Mittal forms as regularization terms, we unveil new families of mirror gradient descent algorithms that can be tailored to the optimization landscape through suitably chosen hyperparameters. 
The adoption of these generalized entropies opens the door to obtaining advantageous properties  such as improved convergence rates, robustness against vanishing/exploding gradients, and inherent flexibility for handling non-Euclidean geometries.
Table~\ref{tab:GEG} summarizes the main results obtained from our study with listing the generalized exponentiated gradient update induced by the deformed exponential functions corresponding to the deformed logarithms used to define the various trace-form entropies.

The theoretical developments presented not only unify additive and  multiplicative gradient update rules via Bregman divergences but also pave the way for designing robust machine learning algorithms that have the ability  to adapt  precisely to the structure of training data distributions via hyperparameters. Future work will investigate broader classes of entropic functions, extending the framework to non-convex and stochastic optimization settings, and applying the proposed approach to practical problems, and perform systematic comparison through computer simulations experiments.


\appendix

\section{Deformed algebra and calculus}

\subsection{$q$-Algebra and Calculus}\label{sec:qalgebra}

In this section we briefly summarize $q$--Algebra \cite{Tsallis1994,Borges2004}:\\

\begin{itemize}
\item

$q$-sum:  $x \oplus_{q} y  = x +  y  +(1-q) x y $\\

\item

neutral element of $q$-sum:   $x \oplus_{q} 0 = 0 \oplus_{q} x  =x $\\

\item

$q$- substraction:  $x \ominus_{q} y  = \frac{x - y}{1+(1-q)y},\;\; y \neq -1/(1-q)$\\

\item

$q$-product:  $x \otimes_{q} y  = \left[x^{1-q} + y^{1-q} +1\right]_+^{1/(1-q)}
 $\\

\item

neutral element of $q$-product:  $x \otimes_{q} 1= 1 \otimes_{q} x  =x $\\

\item

$q$-division:  $x \oslash_{q} y  = x  \otimes_{q} (1/y )=
\left[x^{1-q} - y^{1-q} +1\right]_+^{1/(1-q)} $\\

\item

Inverse of $q$-product  $1 \oslash_{q} x  =\left[ 2-x^{1-q} \right]_+^{1/(1-q)}$.
\end{itemize}

\subsection{$\kappa$--Algebra and Calculus}\label{sec:kappaalgebra}

In this section we briefly summarize $\kappa$--Algebra especially $\kappa$-product properties \cite{kaniadakis_scarfone2002}--\cite{kaniadakis2005}:

\begin{itemize}
\item

$\kappa$-sum:  $x \oplus_{\kappa} y  = x \sqrt{1+\kappa^2 y^2}+  y \sqrt{1+\kappa^2 x^2}$,\\

\item

neutral element of $\kappa$-sum:   $x \oplus_{\kappa} 0 = 0 \oplus_{\kappa} x  =x $,\\

\item

$\kappa$- substraction:  $x \ominus_{\kappa} y  = x \sqrt{1+\kappa^2 y^2}- y \sqrt{1+\kappa^2 x^2}$,\\

\item

 $\kappa$-product:  $ \displaystyle x \otimes_{\kappa} y  =
\exp\left(\frac{1}{\kappa}\text{arsinh}((x^{\kappa} - x^{-\kappa} + y^{\kappa} - y^{-\kappa})/2) \right) $,\\

\item

admits the unity as a neutral element of $\kappa$-product:  $x \otimes_{\kappa} 1= 1 \otimes_{\kappa} x  =x $.\\

\item

$\kappa$-product is commutative: $x \otimes_{\kappa} y = y \otimes_{\kappa} x$\\

\item

$\kappa$-product is associative: $(x \otimes_{\kappa} y) \otimes_{\kappa} z = x \otimes_{\kappa} (y \otimes_{\kappa} z)$\\

\item

 the inverse element of $x$ is  $1/x$, i.e., $x \otimes_{\kappa} (1/x) =1$\\

\item

$\kappa$-division:  $x \oslash_{\kappa} y  = x \otimes_{\kappa} (1/y) =
\exp\left(\frac{1}{\kappa}\text{arsinh}((x^{\kappa} - x^{-\kappa} - y^{\kappa} + y^{-\kappa})/2) \right) $\\

\item

Inverse of $\kappa$-product  $1 \oslash_{\kappa} x  = x^{-1}$.

\end{itemize}

\end{document}